\documentclass{IOS-Book-Article}
\pdfoutput=1

\usepackage{mathptmx}

\usepackage[utf8]{inputenc} 
\RequirePackage{hyperref}
\RequirePackage{hypernat}
\usepackage{enumerate}
\usepackage{enumitem}
\usepackage{graphicx}
\usepackage{url}
\usepackage{amssymb}
\usepackage{enumerate}

\usepackage{algpseudocode, algorithm}

\usepackage{caption}
\usepackage{subcaption}
\usepackage{multirow}

\usepackage{array}
\usepackage{makecell}
\usepackage{graphicx}

\usepackage{1-charigruensenevomcguinness}

\def\hb{\hbox to 10.7 cm{}}

\begin{document}

\pagestyle{headings}
\def\thepage{}
\begin{frontmatter}  
\title{Foundations of Explainable Knowledge-Enabled Systems}

\markboth{}{March 2020\hb}

\author[A]{\fnms{Shruthi} Chari}%
\author[B]{\fnms{Daniel M.} Gruen}
\author[A]{\fnms{Oshani} Seneviratne}
and
\author[A]{\fnms{Deborah} L. McGuinness}

\runningauthor{S. Chari, D. Gruen, O. Seneviratne, D. McGuinness}
\address[A]{Rensselaer Polytechnic Institute, Troy, NY, USA}
\address[B]{IBM Research, Cambridge, MA, USA}

\begin{abstract}
Explainability has been an important goal since the early days of Artificial Intelligence. 
Several approaches for producing explanations have been developed. However, many of these approaches were tightly coupled with the capabilities of the artificial intelligence systems at the time. With the proliferation of AI-enabled systems in sometimes critical settings, there is a need for them to be 
explainable to end-users and decision-makers. We present a historical overview of explainable artificial intelligence systems, with a focus on knowledge-enabled systems, spanning the expert systems, cognitive assistants, semantic applications, and machine learning domains.
Additionally, borrowing from the strengths of past approaches and identifying gaps needed to make explanations user- and context-focused, we propose new definitions for explanations and explainable knowledge-enabled systems.

\end{abstract}

\end{frontmatter}

\section{Introduction} \label{chapter01-introduction}
The growing incorporation of Artificial Intelligence (AI) capabilities in systems across industries and consumer applications, including those that have significant, even life-or-death implications, has led to an increased demand for explainability.  To accept and appropriately apply insights from AI systems, users often require an understanding of how the system arrived at its results. 

Such an understanding can include having a model of how the underlying AI system operates, how it was constructed, and how the data used to develop and train it matches the situations in which it was used.  It can include information about the specific features of the current situation that contributed to the system's determination. It can also include descriptions of the underlying rationales and reasoning paths the system used to arrive at a conclusion, which in turn can be based on observed statistical regularities, models of underlying mechanisms and causal relationships, and temporal patterns. We draw a distinction, between \textit{transparency},\footnote{In this chapter we use quotes for terms that we introduce or for direct quotations from publications, and we use italics to either emphasize terminology from papers or highlight important terms.} by which we mean general information about a system's operation, capabilities, underlying training data, and fairness,  and \textit{explainability}, by which we mean the ability of a system to provide information describing and justifying how a specific result was determined along with the overall context. We build on this notion of explainability and present desired properties for explanations and redefine explanations supporting a user's perspective in Section \ref{chapter01-definition}. 

By their very nature, explanations are user focused; explanations are needed because they provide information that would otherwise be absent that helps a user trust, apply, and maximally benefit from the AI system's operation.  Thus, the need for explanation, and the types of explanations required, are contextual, depending on users, their roles, their prior knowledge, and the situation. For example, a physician recommending a non-standard treatment regimen might want to understand what aspect of the current patient's condition led to an unexpected result, and how the reasoning behind it aligns with scientific knowledge about biological and pharmacological mechanisms.  A patient-facing explanation for the same result may need to include more basic information on the condition and what is unique about the patient's situation.  An explanation aimed at a hospital administrator or insurance coordinator 
may need to include information about potential biases that could lead to a lack of fairness in the recommendation.

Explanations can have deeper value beyond the ``gating" role they play in helping users determine which results should be trusted and applied.  Explanations provided in the above example could contribute to the mental model the physician is constructing of the patient, and of diseases and biological mechanisms in general, that could be valuable in future treatment decisions they make for that patient and others.  Explanations also contribute to the model users are creating of the system itself, by exposing the kinds of information and processing mechanisms the system utilizes.  Norman famously described how users construct mental models of  systems with which they interact across gulfs of execution and interpretation \cite{norman1988psychology}. With AI systems, explanations can help users simultaneously construct models of the system with which they are interacting, and of the underlying domain and situation in which the system is being used.

The importance of explainability is particularly salient with collaborative AI systems meant to work in tandem with human users to augment rather than supplant their skills and capabilities.  A ``Distributed Cognition" approach \cite{hollan2000distributed} is informative here, in which cognition is seen to take place not within the head of any one individual, but rather through the exchange and transformation of representations across multiple actors and artifacts  \cite{hutchins1990technology}. The ability for a system to provide explanations, and respond to queries that reference other information relevant to the situation, expands the range of ways in which the system and human actors can interact.

\subsection{Historical Evolution} \label{chapter01-historicalevolution}
Explainability has been a major goal since the early days of AI. 
In this chapter, we focus on the broad class of knowledge-enabled systems, instead of simply knowledge-based systems.  We include rule-based systems 
as well as hybrid AI systems that may include a wide range of reasoning components including potentially inductive or abductive reasoning as well as the more traditional deductive reasoning.  
As such, we include historical explanation work (e.g., \cite{swartout1991explanations, shortliffe1974mycin, clancey1982neomycin}) and also explanation work aimed more at evolving hybrid AI systems (e.g., \cite{mcguinness2007explaining, seneviratne2018knowledge, gunning2017explainable}). 
The survey includes 
the domains of expert systems, cognitive assistants, Semantic Web \cite{hendler2002integrating}, and, more recently, explanations that work with black-box models, i.e., deep learning models \cite{lecun2015deep}. 
With this background, we will now present a historical perspective on the evolution of explainable AI. 

Many early AI systems took a rule-based, expert system approach. Expert systems (e.g., \cite{shortliffe1974mycin, swartout1991explanations}) were inherently explainable in that they used a set of rules to come to conclusions, so explanations could be generated that provided a detailed or abstracted collection of the rule executions as an explanation of a conclusion.  During the expert system era, much work focused on explaining these systems and their decisions to the end-user. Explanations were broadly intended to address the \textit{Why}, \textit{What}, and \textit{How} aspects of an AI system that produces a result.
Dhaliwal et al. \cite{dhaliwal1996use} provide an overview of these explanation types and state that the \textit{Why} explanations were populated with the justification for a conclusion, the \textit{How} explanations contained a trace of the mechanistic functioning of the system, and, the \textit{What} explanations exposed the system's decision variables involved in the conclusion. Explanations produced by these systems were mainly focused on introducing the rationale behind a system's decision and the way the system works. 
Additionally, while trace-based explanations produced by expert systems captured the \textit{why} and \textit{how} aspects, they typically did not account for the context of a user when they generated explanations.
There were a wide range of expert systems early on. For example, MYCIN \cite{shortliffe2012computer} was an early expert system that supported medical diagnosis using a rule-based inference engine and included a trace-based explanation component.

Today, with the availability of vast volumes of data, deep learning algorithms are being widely used. However, these models are largely uninterpretable,
and a significant focus of explainable AI research
(e.g., DARPA XAI Report 2017 \cite{gunning2017explainable}) is focused on explaining the underlying mechanisms of these black-box models. In our opinion, gaining transparency into the black boxes can be useful, and it may decrease 
the ``unintelligibility aspect"
\cite{lou2012intelligible}. However, it is not enough to
provide \textit{personalized, tailored, and trustworthy} \cite{holzinger2017we} explanations to consumers of AI models. Additionally, machine learning (ML) models often
output a score or probability as predictions. While the number may be useful to understand some level of confidence, a single number lacks context, and thus is often inadequate without additional information.
Semantic Web representation and reasoning work is well suited to help here.  Standards for representing terminology, e.g., RDF and OWL \cite{mcguinness2004owl}, as well as representing provenance (e.g., PROV-O \cite{lebo2013prov} and nanopublications \cite{groth2010anatomy}), have emerged and can be used to encode information along with its provenance and a system's reasoning provenance, and this may be used to augment explanations.
The inclusion of provenance into the underlying representation and thus potentially into the
explanations partially addresses the \textit{What} and \textit{Why} aspects of the reasoning behind presenting explanations to consumers. 

Recently, researchers have acknowledged an increasing
need to include explainability modules into AI systems. As a consequence, several survey papers \cite{mittelstadt2019explaining, biran2017explanation, holzinger2017we} have highlighted
past noteworthy efforts in explainable AI.\footnote{We choose to use the explainable AI phrasing for explainability efforts in AI as XAI 
has come to be associated with the DARPA explainable AI (XAI) program. Our focus is broader than the program's focus on explaining and interpreting black box ML methods from a cognitive perspective.} These survey papers emphasize the fact that different situations, users, and contexts demand different kinds of explanation \cite{arya2019one}. Different AI systems are geared toward addressing an explanation type (or rarely, a combination), e.g., expert systems typically provide trace-based explanations, deep learning models can be leveraged to offer contrasting explanations, etc. We believe that the next-generation AI systems need to go beyond the \textit{Why, What, How} aspects and produce explanations that additionally prioritize issues related to the setting, users' understanding, and contexts. At a minimum, these AI systems need to include a provenance component to support trust and provide users with tools that can access a reasoning trace to further explore or serve as a means of understanding. In Section \ref{chapter01-Approaches}, we will review a few of these past approaches mentioned in this section.

\subsection{Shift and Current Focus of Explainable AI} \label{chapter01-shift}
The evolution of AI systems has been heavily influenced
by the availability of resources, computing power and data. As previously
stated, AI has moved from primarily using rule-based expert systems to using
ML methods, and, sometimes, hybrid methods.  With
these changes, there has been a shift in the focus of explainability, due to the new challenges of the interpretability of complex ML models.
The initial explanation focus on working from system traces to provide a notion of what was done has expanded to including a focus on including a notion of interpretability of an underlying ML model.  This \textit{interpretable} ML work may be a first generation of explanation of ML, but as we will expand on below, more is needed. 
Additionally, in the Semantic Web \cite{hendler2002integrating}, and more generally, in knowledge representation-based applications, the focus has expanded from traditional \textit{What} explanations to include explanations
addressing information attribution and provenance aspects. The motivations for those expansions include improving
the trustworthiness of information being represented in knowledge graphs (KGs), and 
further, to provide more context for users as they are deciding how to use the information in analysis applications.
Further, AI models are now being employed in user-facing settings where there is a need for personalized conclusions. Hence, there is a need to rethink explanations produced by AI systems from a user perspective and include components to educate users, 
align with their cognitive model,
help them trust the system, provide relevant information, and tailor suggestions to a user's contexts \cite{mittelstadt2019explaining, biran2017explanation}. Borrowing strengths from explanations provided by past approaches, we will attempt to present, synthesize, and
refine a definition
of explanation and explainable knowledge-enabled systems with an acknowledgment of desired explanation properties that fit today's settings.
\section{Terminology}
\label{chapter01-definition}
Several researchers have proposed comprehensive definitions of explanations \cite{swartout1993explanation, doshi2017accountability, mcguinness2007explaining, gilpin2018explaining} and have presented explanation components that they deem necessary to satisfy either their work or the domains
where they hope the explanations will be useful. However, with a shift of focus in AI we feel the need to revisit the work on defining explanation as we consider what is desirable in 
next-generation ``explainable knowledge-enabled systems." In this section, we list desirable properties (Section \ref{chapter01-desirableprops}) for both explanations and explainable knowledge-enabled systems that generate these explanations, and use these properties as a basis to provide definitions.

\subsection{Desirable Properties}
\label{chapter01-desirableprops}

\subsubsection{Explanations}

As a part of our list of desirable properties for explanations, we present properties, such as, \textit{improving user appeal, and achieving user understandability}, that have been explored as explanation components in the past, and that will be useful in designing explanations suitable for the end-user. In addition, we propose including a higher priority on the inclusion of features, such as, \textit{provenance} and \textit{adapting to user's context} that will have a renewed focus in making explanations user-centric and in mitigating the unintelligibility aspect of current ML methods. 
We are aligning with others who have called for a greater user focus in explanations \cite{miller2019explanation, lim2009and, ribera2019can}. 

\begin{itemize}
    \item \textbf{Be understandable:} Borrowing from desired properties of explanations stated by Swartout and Moore \cite{swartout1993explanation}, we highlight that for explanations to be \textit{understandable} by the user, the explanations should use terminology familiar to the user.  If terminology is potentially unfamiliar, then we also suggest that capabilities be included for obtaining definitions of terms, thereby \textit{educating} users. 
    Understandability has the potential to be significantly increased if the AI system incorporates \textit{user feedback} and a model of \textit{user context}.
     \item \textbf{Include provenance:} \textit{ Provenance} is a property of explanations that has either been absent in some past descriptions of explanations \cite{swartout1993explanation}, or has not had the emphasis that it deserves now. 
    As systems expand to include more diverse content the need for capturing provenance increases.
   Explanations need to include \textit{provenance}
that includes information about the domain knowledge utilized by the system,
along with the methods used to obtain that knowledge.  
We borrow from the ``counterfactual faithfulness" idea proposed by \cite{doshi2017accountability}, and argue that, as part of the \textit{provenance} components, explanations need to carry \textit{causal information} about the conclusion, if present in domain literature or supported by expert knowledge.
    \item \textbf{Appeal to user:} Paraphrasing Swartout and Moore \cite{swartout1993explanation}, we note that explanations need to be \textit{rich, coherent, and appeal to the user}.We propose that explanations need to expose facts that the user finds \textit{resourceful and sufficient for further exploration}. A \textit{resourceful} 
    explanation contains enough granular content and evidence to appeal to the user's mental cognition and current needs. A \textit{sufficient} explanation contains content that the user requires to carry out their tasks. A subtlety in generating explanations that appeal to the user would be to tailor the explanation length to the user's needs and preferences, i.e., to avoid lengthy explanations with content that might not be useful to the user or that they already understand. Further, we acknowledge that the resourcefulness and sufficiency aspects of explanations might be hard to measure in real-time. However, we suggest that explainable knowledge-enabled systems should be designed after an analysis of user requirements and utilize techniques to employ dynamic and static evaluation strategies to help realize these goals. More specifically, dynamic strategies could involve interactive mechanisms, such as the delivery of persuasive messages used by Maimone et al. \cite{maimone2018perkapp}, and static evaluation strategies could include user surveys conducted to evaluate the effectiveness of the systems, such as the one by Glass et al. \cite{glass2008toward}. 
    \item \textbf{Adapt to users' context:} 
Besides being user-centric, explanations need to be tailored to the user's \textit{current scenario and context}. Explainable AI systems not only need to leverage information about the user (as may have been captured in a user profile \cite{rich1989stereotypes, sugiyama2004adaptive}), 
but they also 
need to identify the user's intent and adapt the \textit{explanation form} to connect to the user's mental model and align with the user's intent. 
For example, an explanation may include a contrastive hypothesis that relates to the user's intent or statistical evidence to provide more support to enhance a user's belief. 
In a later chapter, ``Directions for Explainable Knowledge-enabled Systems,” in this book, we present different explanation types and their various focii that would allow AI systems to generate diverse explanations. 
\end{itemize}

Overall, explanations should serve beyond their original aim to teach \cite{swartout1991explanations},
and provide \textit{trustworthy, transparent, unambiguous accounts} of automated tasks to end-users.

\subsubsection{Explainable Knowledge-Enabled Systems} \label{chapter01:secexplainablekbfeatures}
While many have attempted to define explanations (e.g., \cite{swartout1993explanation, doshi2017towards}), additional efforts have attempted to improve the generation of explanations (e.g.,  \cite{mcguinness2007explaining, glass2008toward, air-policy-language}) and tackle various aspects of explainability (e.g., \cite{pearl2009causality, mittelstadt2019explaining}). To begin to address the need of building explainable, knowledge-enabled AI systems, we present a list of desirable properties  from the synthesis of our literature review of past explanation work. Our review primarily spans knowledge representation in expert systems \cite{swartout1993explanation}, provenance and reasoning efforts in the Semantic Web \cite{groth2010anatomy}, user task-processing workflows in cognitive assistants \cite{mcguinness2007explaining, mcguinness2004explaining}, and efforts to reduce unintelligibility in the ML domain \cite{gunning2017explainable, gilpin2018explaining, arya2019one}. Additionally, we analyzed explanation requirements from current literature, answering an increased need for \textit{user-comprehensibility} \cite{lecue2019role}, \textit{accountability} \cite{doshi2017towards} and \textit{user-focus} \cite{mittelstadt2019explaining}. In our literature review in Section \ref{chapter01-Approaches} we will highlight approaches that exhibit these properties.

\begin{itemize}
    \item \textbf{Modularity}: A modular design, such as, the one proposed by Swartout and Moore \cite{swartout1991explanations}, is desirable, as it would allow systems to adapt models and functioning to users' requirements and scenarios. This property would also allow for the AI system to include explanation facilities that tap into various modules to expose information requested by and conducive to the user's needs.
    \item \textbf{Interpretability}:
    Borrowing from Mittlestadt et al. \cite{mittelstadt2019explaining} and Hasan and Gandon \cite{hasan2012explanation}, we believe
    that the interpretability of explainable knowledge-enabled systems enables them to be transparent, lending to the ability to provide trace-based accounts of their working. Additionally, we utilize Gilpin et al's definition of interpretability as a  ``science of comprehending what a model did." However, if the models used in the system are not interpretable, we propose that they should consider including proxy methods to be interpretable, for example, utilizing linear proxy models proposed by Gilpin et al. \cite{gilpin2018explaining} that serve as a simplified proxy of the full model. 
    \item \textbf{Support provenance:} Paraphrasing from the explanation requirements suggested by Hasan and Gandon \cite{hasan2012explanation}, we agree that explainable knowledge-enabled systems should store the provenance of the information that their models rely on beyond just metadata. We believe that the inclusion of provenance aids AI systems in generating \textit{resourceful and sufficient} explanations for users, providing them with resources for further exploration.  
    \item \textbf{Adapt to user's needs:} We propose that AI systems need to be \textit{adaptive} and \textit{interactive}, adapting their functioning and explanation generation capabilities to suit the user's requests and contexts. To this end, and to provide tailored explanations, Ribera and Lapedriza \cite{ribera2019can} have identified user categories (domain experts, AI researchers, and lay users) and presented their contrasting demands from an AI system. Further, the ability to be adaptive would be enhanced by a modular design, as suggested earlier, and would aid the system in generating explanations in various forms to suit the user's understanding and their needs. 
    \item \textbf{Include explanation facilities:} Inspired by McGuinness et al.'s cognitive assistants explanation frameworks \cite{mcguinness2007explaining, mcguinness2004explaining}, we propose that the design of the explanation facilities should be addressed early and in detail in the design phase, to ensure that the AI system is capable of supporting the requirements of the explanation facilities within its design. 
Explanation facilities could constitute a wide-range of user-facing interfaces, such as, dialogue systems, visualizations, and feedback systems that the user interacts with and provides feedback to the AI system about the explanations generated or a need for further clarifications. Hence, since explanation facilities would require additional information, such as, provenance, and would need the system to incorporate feedback and adapt to context, we recommend that their design be coupled with the AI system design. 
    \item \textbf{Include/Access a knowledge store}:  We recommend that explainable knowledge-enabled systems store the \textit{domain knowledge} they rely on, the \textit{user's mental model} they appeal to, and the \textit{explanation components} they are generating. Additionally, we relax the inclusion of knowledge in that an AI system might provide access to a knowledge store - as the system may host it, or it may use some other system's hosting and contribute to and access that store. By knowledge store, we refer to data storage mechanisms (KGs or semantic representations are preferred) that can store knowledge of various forms spanning categories such as background knowledge, domain knowledge, etc. 
    \item \textbf{Support compliance and obligation checks:} 
    In addition to hosting/accessing knowledge stores, we recommend that explainable knowledge-enabled systems store an encoding set of expert knowledge in their field of application. These encodings
    should be sufficient to determine if the system complies with the standards and practices in that field. Additionally, we also recommend that explainable knowledge-enabled systems attempt to adhere to standards for the proposed explainable AI models, such as \cite{gilpin2018explaining, arya2019one}. Furthermore, we suggest that compliance and obligation checks be evaluated on the system post-construction.
\end{itemize}  

\subsection{Definitions} \label{chapter01-definitions}
Having identified desirable properties for explanations and explainable knowledge-enabled systems, will now provide a set of definitions leveraging our
review of the explanation literature and our analysis of the current AI landscape. Our goal is to reflect the needs of explainable AI in current times and provide a summary of the desirable properties to achieve better explainability. 
\subsubsection{Explanation}
We define an explanation in the computational world as, ``an account of the system, its workings, the  \textit{implicit and explicit} knowledge used in its reasoning processes and the specific decision, that is \textit{sensitive} to the end-user's \textit{understanding, context, and current needs}." 
\subsubsection{Explainable Knowledge-Enabled Systems}
We define ``explainable knowledge-enabled systems" to be, ``\textit AI systems that include a representation of the domain knowledge in the field of application, have mechanisms to incorporate the \textit{users' context}, are \textit{interpretable}, and host \textit{explanation facilities} that generate \textit{user-comprehensible, context-aware, and provenance-enabled} explanations of the mechanistic functioning of the AI system and the knowledge used."
\section{Approaches} \label{chapter01-Approaches}
We present past approaches that have addressed various aspects of explainability related to trust, transparency, provenance and interpretability. To the extent possible, we group publications by technical domain: knowledge-based systems, Semantic Web applications, cognitive assistants, and ML systems, in an attempt to show the progression of methods within those domains. In
Section \ref{chapter01-explainablekbs}, we consider work from the 1970s-1990s that sought to utilize the trace explainability strengths of rule-based systems to explain the process used to arrive at decisions. 
In Section \ref{chapter01-semanticweb}, we review provenance and explanation modeling efforts and posit them as contributors to the development of trustworthy and explainable semantic applications. 
In Section \ref{chapter01-cogassistants}, we focus on efforts to explain task-based workflows in personal assistants and intelligent tutoring settings. We end with a review of papers that improve the interpretability and trust aspects of ML methods in Section \ref{chapter01-explainable AIworld}. While each of these vast domains has large volumes of published literature, we restrict ourselves to seminal work on explainability in the domain or publications 
that have introduced novel techniques to tackle different aspects of explainability. As a conclusion of each domain subsection, we provide a brief summary 
of the methods utilized to address explainability and describe any lessons applicable for the development of future explainable AI methods. 

\begin{table}[ht!]
 \caption{Foundational explanation approaches and 
 desired features of explainable knowledge-enabled systems}
 \label{tab01:approachesevaluation}
\centering
\resizebox{\textwidth}{!}{%
\begin{tabular}{|l|l|l|p{0.5cm}|p{0.5cm}|p{0.5cm}|p{0.5cm}|p{0.5cm}|p{0.5cm}|p{0.5cm}|}
\hline
\textbf{Year} &
  \textbf{System} & \textbf{Application} &
  \begin{tabular}[c]{@{}l@{}}{\rotatebox{90}{\textbf{Knowledge Store}}}\end{tabular} &
  \begin{tabular}[c]{@{}l@{}}\rotatebox{90}{\textbf{Interpretable}}\end{tabular} &
  \begin{tabular}[c]{@{}l@{}}\rotatebox{90}{\textbf{Explanation Facilities}}\end{tabular} &
  \begin{tabular}[c]{@{}l@{}}\rotatebox{90}{\textbf{Modular}}\end{tabular} &
  \begin{tabular}[c]{@{}l@{}}\rotatebox{90}{\textbf{Adaptive}}\end{tabular} &
  \begin{tabular}[c]{@{}l@{}}\rotatebox{90}{\textbf{Support Provenance}}\end{tabular} &
  \begin{tabular}[c]{@{}l@{}}\rotatebox{90}{\textbf{Compliance Checks}}\end{tabular} \\ \hline
1977 & MYCIN \cite{shortliffe1974mycin} & Healthcare                                                                                           & \ding{51}
  & \ding{51}
 &  &  &  &  &  \ding{51} \\ \hline
1982 & NEOMYCIN   \cite{clancey1982neomycin} & Healthcare &                                                     \ding{51}
  & \ding{51}
  & \ding{51} & \ding{51} &  & \ding{51}
  & \ding{51} \\ \hline
1989 & CLASSIC    \cite{borgida1989classic} & General Purpose                                                                                        & \ding{51}
 & \ding{51}
 & \ding{51}
 & \ding{51}
 & \ding{51}
 &  & \ding{51}  \\ \hline
     1991 & \begin{tabular}[c]{@{}l@{}}Explainable Expert \\ System (EES) \cite{swartout1991explanations} \end{tabular}   & Program Advisor &                        \ding{51}
&  \ding{51} & \ding{51}  & \ding{51}
 &  &  & \ding{51} \\ \hline
     2004 & \begin{tabular}[c]{@{}l@{}}Inference Web \\  Framework \cite{mcguinness2004explaining} \end{tabular}                                & General Purpose & \ding{51} & \ding{51}  & \ding{51} & \ding{51} & \ding{51}  & \ding{51} & \ding{51} 
     \\ \hline
     2005 & Disciple-LTA \cite{tecuci2005personal} & Military & \ding{51}  & \ding{51} & \ding{51} & \ding{51}  & \ding{51} & \ding{51} & \ding{51} \\ \hline
     2007 & \begin{tabular}[c]{@{}l@{}}Integrated Cognitive \\ Explanation \\ Environment (ICEE) \cite{mcguinness2007explaining} \end{tabular}    & Office Assistant & \ding{51} & \ding{51}  & \ding{51} & \ding{51} & \ding{51}  & \ding{51}   & \ding{51} \\ \hline
     2009 & \begin{tabular}[c]{@{}l@{}}Automated Policy \\ Reasoning \cite{air-policy-language} \end{tabular}                                & Judicial & \ding{51} & \ding{51} & \ding{51}  &  & \ding{51}  & \ding{51}  &  \ding{51} \\ \hline
      2016 & \begin{tabular}[c]{@{}l@{}}Local Interpretable \\ Model-agnostic\\ Explanations (LIME) \cite{ribeiro2016should} \end{tabular} & General Purpose & & \ding{51} & \ding{51} & \ding{51} &  &  &  \\ \hline
     2017 & ReDrugs \cite{mccusker2017finding}                                                                                   & Medicine & \ding{51} & \ding{51} &  &  &  & \ding{51} & \ding{51} \\ \hline
    %  
    % 2018 & \begin{tabular}[c]{@{}l@{}}Breast Cancer \\ Staging Application\end{tabular}                       & \ding{51} & \ding{51} &  &  &  & \ding{51} &   \\ \hline
     \begin{tabular}[c]{@{}l@{}}2017 \\ (ongoing)\end{tabular} & \begin{tabular}[c]{@{}l@{}}Common Ground\\  Learning and \\ Explanation (COGLE) \cite{gunning2019darpa} \end{tabular}  & Unmanned Aircrafts & \ding{51} &  & \ding{51} & \ding{51} & \ding{51} &  & \ding{51} \\ \hline
\end{tabular}%
}
\end{table}

Table \ref{tab01:approachesevaluation} contains an evaluation of the foundational AI systems, reviewed against the criteria we defined for explainable knowledge-enabled systems (Section \ref{chapter01-desirableprops}). 
The chronological order allows us to view trends in explainability over the years and also helps expose shifts in the areas of focus and strengths of the class of approaches. We observe that explanations were well-explored as a topic of interest in the AI community from the early 1990s - mid-2000s. We note that, even within the expert systems era, the AI architecture evolved from simply generating trace-based accounts of decisions to including modular explanation facilities (\cite{borgida1989classic, swartout1991explanations, clancey1982neomycin}) that sometimes could produce provenance-enabled (\cite{clancey1982neomycin}), adaptive and user-customizable (\cite{patel1991classic}) explanations. Additionally, observe that, among other classes of approaches in our review, explanations have been best established in cognitive assistants, which also have the most direct impact on human decision-making capabilities. However, we notice that, with more recent systems in the Semantic Web and ML domains, there has been a shift in explainability from building explanation facilities to minimally ensuring that AI models are interpretable (\cite{ribeiro2016should} and support provenance \cite{mccusker2017broad} for further tracing. Further, most AI systems in our review (\cite{shortliffe1974mycin, clancey1982neomycin, borgida1989classic,  swartout1991explanations, mcguinness2004explaining, mcguinness2007explaining, tecuci2005personal, mccusker2017broad}) satisfy the `Compliance Checks' criteria by leveraging logical rule-based or other deductive reasoners to check or enforce compliance.
Also, systems such as the Disciple-LTA and Common Ground Learning and Explanation (COGLE) deployed in critical settings of military and aviation, respectively, have features to allow both expert and lay users to provide feedback about the system's explanations and outputs. Hence, indicating that system supported features are partially driven by the domain of application. Finally, while our evaluation was conducted on a carefully selected set of approaches, our findings on explainability trends are in-line with a larger, systematic review conducted by Nunes and Jannach \cite{nunes2017systematic}, who noted that explainability was best explored in the expert systems and cognitive assistants domains.

\subsection{Knowledge-based systems} \label{chapter01-explainablekbs}
The 80s decade saw the rise of knowledge-based and expert systems, that were designed to assist humans where human resources were limited \cite{dhaliwal1996use}. 
Expert systems and knowledge-based systems both contained an encoding of knowledge. More specifically, in the case of expert systems, the knowledge encoded was that of expert's knowledge, typically in the form of rules. In our review, we will not make distinctions between these two classes of systems and will focus on identifying the explainability components of these systems.  
 From an implementation perspective, both of these systems required the engineering and encoding of multiple rules to support inference. This reliance on rules made these systems inherently explainable, as one could trace back the rules to identify the factors that lead to a conclusion. Subsequently, researchers have introduced different types of explanations \cite{clancey1982neomycin, shortliffe1974mycin}, and approaches to improve explanation generation \cite{borgida1989classic}, and to introduce more granular content into explanations generated by these systems \cite{swartout1991explanations}. 

\subsubsection{Early Expert Systems: MYCIN and NEOMYCIN}
The MYCIN \cite{shortliffe1974mycin, shortliffe2012computer} paper was one of the first to introduce computer-based explanations, and, is regarded as a foundational and seminal work. The goal of the MYCIN system was to identify highly probable carriers of infectious diseases, and suggest treatments for the diseases. The system provided explanations by exposing the inference trace that lead to a decision. The system was able to trace back and expose the reasoning, that served as justifications of decisions.
In particular, MYCIN provided \textit{Why} and \textit{How} explanations \cite{dhaliwal1996use}. The \textit{Why} explanations included facts and task-based information to address a user's queries. The \textit{How} explanations explained the manner and trace in which the system generated the conclusion.

To enhance the \textit{Why} and \textit{How} explanations, a descendant of MYCIN - NEOMYCIN \cite{clancey1982neomycin} produced strategic explanations comprised of meta-knowledge and the problem-solving strategies to adapt the MYCIN system to a teaching setting. NEOMYCIN built on MYCIN's inability to explain beyond the expert knowledge known to the system and 
added a component that leveraged explicit encodings of
problem-solving strategies used to generate the medical knowledge for use in its explanations. To this end, the NEOMYCIN system used a meta-strategy to decide what portion of the rules to invoke from data sources, including an etiological taxonomy, disease knowledge, and causal associations. The metastrategy contained rules that a human would use to undertake tasks such as building hypothesis, pursuing them, identifying problems, etc. In essence, the NEOMYCIN system attempted to mimic human decision-solving, where one would eliminate a hypothesis based on the search space, and not by merely navigating the knowledge (``bridge concepts"  \cite{clancey1982neomycin}) that the system already holds. Further, NEOMYCIN introduced the idea of separating knowledge to make the system more accessible, which was further adopted by Moore and Swartout in their Explainable Expert System \cite{swartout1991explanations} effort, discussed later in this section.
While the strategic explanations generated by NEOMYCIN are desirable, they might be onerous for user consumption due to a surplus of details. 
\subsubsection{Explainable Description Logics: CLASSIC} \label{chapter01-sec:CLASSIC}
McGuinness and Borgida took an approach to explanation where each of the inferences that the underlying logical reasoning system could execute had a declarative explanation description and those individual explanation components could be used to build simple, complex, abstracted, or otherwise customized explanations \cite{mcguinness1995explaining}.  Additionally, every expert rule that a knowledge-based system builder encoded in the system included a structured component that could be used to explain when that rule was used.   These explanation ``breadcrumbs" could then be used to assemble explanations when a user's actions triggered the execution of a rule.

The authors implemented their approach in the CLASSIC knowledge representation system, a description-logic-based language that provided a framework ``to define structured concepts and make assertions about individuals in a knowledge base" \cite{patel1991classic}.  The complete set of foundational inference rules that could be explained for the underlying description logic reasoner was also available for reuse in other systems \cite{mcguinness1996explaining}. 
This style of encoding axioms for every inference that a system could execute was also leveraged in the axiomatic semantics for other predecessors to today's description logic-based recommended language for encoding ontologies on the web: OWL \cite{mcguinness2004owl}. The axioms for RDF, RDFS, and DAML+OIL were described in W3C Note\footnote{DAML+OIL axioms note link: \url{https://www.w3.org/TR/daml+oil-axioms }}
and then were used in a number of different reasoners to provide trace-based explanation capabilities.

\subsubsection{Explainable Expert System}
Moore and Swartout coined the term `Explainable Expert Systems' (EES) in their widely-cited work \cite{swartout1991explanations}. 
The EES framework that aimed to provide explanations and was tested in a Program Enhancement Advisor setting. The explanations generated by the EES system borrowed from and had components of various knowledge sources including domain, problem-solving and system terminology.
Further, the design of their EES system supported the generation of the various components of the explanations and were made of knowledge bases, a program writer, an explanation generator, an interpreter, and an execution trace. The EES system used 
a planning algorithm, wherein goals are reformulated if no viable match is found in the domain knowledge. The reformulation of goals was achieved by the
representation of the domain knowledge into a concept hierarchy, via a language, such as, KL-ONE \cite{brachman1989overview}. The EES framework was interactive in nature, and goals were reformulated based on user dialogue with the system. Additionally, users' queries were used as a cue to interleave domain and problem-solving knowledge traces into their explanations.

\subsubsection{Summary}
The explainable knowledge-based systems 
that we discussed introduced several types of explanations, including \textit{Why, How, and Strategic} explanations (described earlier in Section \ref{chapter01-historicalevolution}). However, their reliance on encoding a large rule base makes them difficult to scale and extremely human-intensive to maintain. Today, we see the semi-automatic generation of rules and knowledge-base population via natural language processing and ontology-enabled extraction techniques. Many learnings from knowledge-based systems
have been reused
and expanded in the Semantic Web, as will be illustrated in Section \ref{chapter01-semanticweb}.
\subsection{Semantic Web}
\label{chapter01-semanticweb}
The creation of the World Wide Web (WWW) \cite{berners1994world} made it possible to create content online and make existing content available online in digital formats. In their seminal paper, Berners-Lee, Hendler, and Lassila \cite{berners2001semantic} state that the Semantic Web was intended to unify content being published online through tagging content with unique identifiers, or Uniform Resource Identifiers (URIs), representing the content utilizing well formed definitions from taxonomies and ontologies,
 and borrowing from the knowledge representation world to utilize structuring mechanisms for data. While these properties are desirable and necessary to enable data sharing and achieve a semantic understanding of digital content, they are not sufficient to make the content explainable to a broad range of users. However, the Semantic Web community has tackled the provenance aspect and trace-based aspects of explainability and developed several provisions to both include provenance in the semantic representation \cite{groth2010anatomy, lebo2013prov} and to supporting reasoning mechanisms, \cite{hogan2009scalable} to generate traces. As a direct consequence of the Semantic Web, the textual content is more accessible in knowledge graphs (KGs) via semantic representations \cite{ehrlinger2016towards}. Additionally, KG provisions have made it possible to provide justifications and provenance to suggestions. In this section, we will review some provenance encoding efforts and explainable semantic applications. 

\subsubsection{Provenance modeling efforts}
There have been two somewhat recent foundational
provenance efforts that paved the way for provenance-aware applications, namely the World Wide Web Consortium's work on a recommended standard for provenance on the web (PROV) with its
associated encoding as an ontology
PROV-O
 \cite{lebo2013prov} and nanopublications \cite{groth2010anatomy}.
Nanopublications provide a structure to associate triple statements with their provenance. In general, provenance is essential as it 
encodes information that can be used to explore where information came from and this information can be used to build trust in applications when they use this information to expose the evidence behind their recommendations. 
\subsubsection{Nanopublications}
Nanopublications
were conceived to help disambiguate and represent the context for scientific statements that were extracted from textual corpora and made available as triples. The authors identified that contextual information present in a document was imperative to understand a statement in relation to the full document.
Hence, they designed nanopublications that provided a mechanism to associate metadata or annotations with statements. The schema of nanopublications has evolved over the years. In its current state,\footnote{Nanopublication Guidelines: \url{http://nanopub.org/guidelines/working_draft/}} nanopublications are composed of three named graph components, Assertions, Publication Information, and Provenance. 
The Publication Information graph stores metadata information about the creation of the content, or how it came to be,
such as, the date of creation, author, etc. The Provenance graph contains metadata, such as, citation information. The assertion graph contains one or more subject-predicate-object statements with domain content.

Kuhn et al. \cite{kuhn2013broadening} have proposed an Atomic, Independent, Declarative, and Absolute (AIDA) framework to encode atomic and indisputable assertion statements. They describe a metananopublication world in which nanopublications can be created from other nanopublications via different channels, for example, from authors creating content from scientific results, and from data mining algorithms generating nanopublications from existing unstructured data sources. Essentially through the metananopublications concept, the authors highlight that provenance can be interleaved and chained, to reflect the real world where multiple entities depend on each other at various levels of granularities.

\subsubsection{The Provenance Ontology (PROV-O)}
The PROV-O ontology \cite{lebo2013prov} provides a formal mechanism to support comprehensive modeling of
the provenance of digital objects. In their ontology, they support three primary forms of provenance contributors, agent-centered, object-centered, and process-centered forms.
In PROV-O \footnote{PROV-O ontology W3C note: \url{https://www.w3.org/TR/prov-o/}}, provenance is modeled via three simple class types, i.e., `entities'\footnote{Classes and properties are referred to by their label, and are enclosed within single quotes} which are generated by activities, and `entities' and `activities' that are `associated with' and `attributed to' agents, respectively. In the W3C note, the editors showcase the adequacy of the PROV-O ontology in modeling a use case where a blogger is exploring the provenance chain of a newspaper article while finding out who compiled the chart included in the article. The use case also illustrates that provenance needs to be modeled comprehensively to ensure that users have a complete understanding of the information they are viewing. 

There have been ontology alignment efforts on the PROV-O ontology to enhance usability and increase interoperability. These efforts include alignment of PROV-O with standard ontologies, such as, the TIME ontology, Semantic Sensor Network Ontology (SSN), and the Basic Formal Ontology (BFO). The PROV-O ontology has also served as a foundational ontology for several other provenance ontologies (e.g., Provenance for Clinical and Healthcare Research (ProvCare \cite{valdez2017provcare}) and Guideline Provenance Ontology (G-Prov \cite{agun2019gprov})) that support provenance modeling in specific use cases with different levels of granularity.

\subsubsection{Provenance and Related Semantic Knowledge Graphs}
The Semantic Web community also allows for different alternatives of representing that information based on granularity and content needs such as named graphs, \footnote{Named Graphs: \url{https://www.w3.org/2009/07/NamedGraph.html}} reification, \footnote{Reification: \url{https://www.w3.org/TR/rdf-primer/}} etc., and there exist cross-domain open source KGs that host somewhat comparably rich provenance (e.g., Wikidata, \footnote{Wikidata: \url{https://www.wikidata.org/wiki/Wikidata:Main_Page}} WebIsALOD \cite{hertling2017webisalod}).
Additionally, while we believe that provenance modeling is crucial to provide high-quality, trustworthy information to consumers, we acknowledge that it is not sufficient to capture user context or to personalize results. 
Recently, there has been an emergence of KGs that encode contextual and personal information \cite{gyrard2018personalized, maimone2018perkapp}, lending to the personalizing of semantic applications that are enabled by these KGs. Gyrard et al. \cite{gyrard2018personalized} described the components of a personalized healthcare knowledge graph (PHKG) that are needed to monitor user health to help users combat chronic diseases, such as, asthma and obesity. In a similar effort, Maimone et al. developed Perkapp \cite{maimone2018perkapp}, a persuasive system that monitors people's lifestyles and persuades them to make healthier choices and stay on track. 
Their persuasive, knowledge-based system architecture contains a set of expert-generated rules and outputs persuasive context-aware messages to users based on their 
adherence to the rules.

\subsubsection{Reasoning Efforts}

We now present a selective overview of the reasoning efforts. We briefly introduced RDFS reasoning efforts in Section \ref{chapter01-explainablekbs}. RDFS reasoning results in justifications or trace-based accounts of \textit{Why} a conclusion was made by the system, based on which rule fired. However, these justifications can be overwhelming for human consumption. To address this, Horridge et al. \cite{horridge2008laconic} proposed laconic and precise justifications that do not 1. conceal detail, 2. expose axioms that are relevant to the justification, and 3. are atomic, in that multiple fine-grained cores can be highlighted. Besides the laconic justification effort, there have been other efforts to improve explainability of justifications and we discuss one such effort, the AIR (Accountability in RDF or AMORD\footnote{AMORD (A Miracle Of Rare Device) is an explanation system developed for MIT scheme in the 1970's}~\cite{kleer1977amord} in RDF) language.
\subsubsection{Explanations for Automated Policy Reasoning}
 AIR language that had a broader focus on modeling explanations serving to explain inference traces from policy reasoning. AIR 
is a Semantic Web-based rule language focused on generating and tracking explanation for inferences and actions~\cite{air-policy-language}. The Massachusetts Institute of Technology (MIT) Decentralized Information Group developed the AIR language, as an extension to N3Logic~\cite{berners2008n3logic} to support accountable privacy protection in Web-based information systems conforming to Linked Data principles. Accountability and privacy protection are enabled through auditable trace-based explanations. AIR supports \emph{Linked Rules}, which can be combined and reused like Linked Data. Additionally, AIR explanations can be used for further reasoning.

\begin{figure}[ht!]\centering
\includegraphics[width=1.0\textwidth]{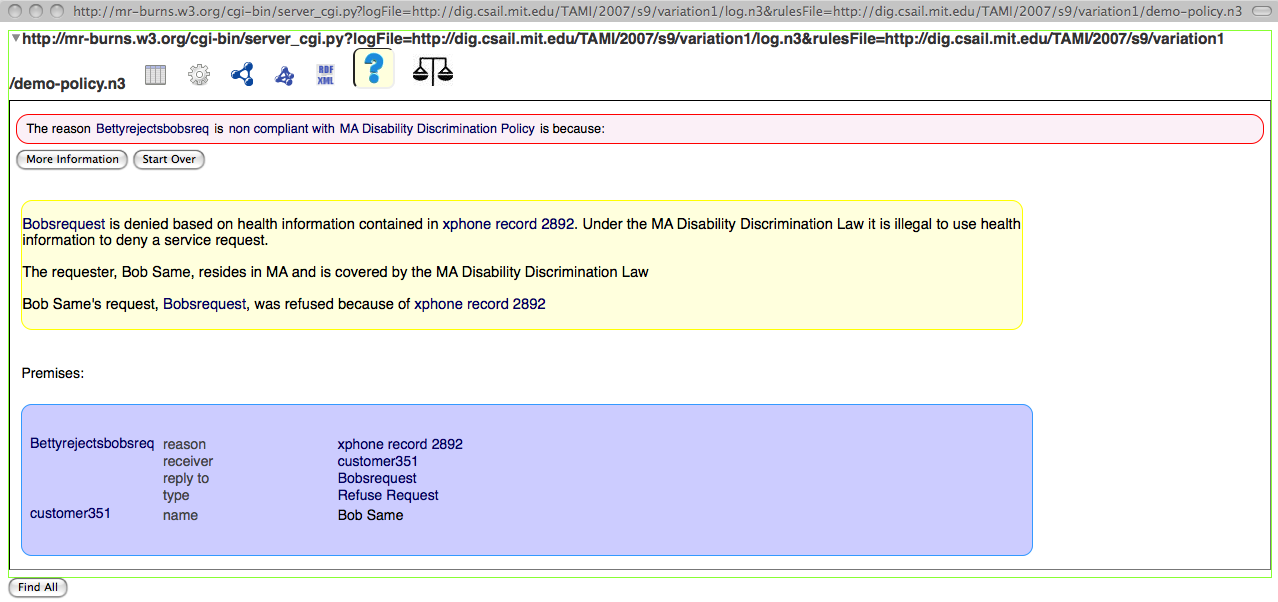}
\caption[tabulatorwhypane]{AIR Justification or Explanations View: Once a user clicks the ``Why?" button, they will see a description appear in the ``Because" box, and the premises that support the justification appear in the ``Premises" box. When the user clicks the ``More Information" button, the descriptions corresponding to outer rules in the proof tree will be appended to the ``Because" box, and the ``Premises" box is overwritten with the corresponding set of premises in the proof tree. When all the descriptions in the proof tree have been traversed, the message ``No more information is available from the reasoner" will be displayed in the ``Premises" box. At any given time, this proof exploration can be restarted by clicking the ``Start Over" button. [Image taken from website \protect\footnotemark]}
\label{chapter01-tabulator_why_pane}
\end{figure}

AIR provides two independent ontologies. One ontology allows the specification of AIR rules,\footnote{AIR rules ontology: \url{http://dig.csail.mit.edu/TAMI/2007/amord/air}} and the other one allows describing justifications.\footnote{AIR justifications ontology: \url{http://dig.csail.mit.edu/2009/AIR/airjustification.n3}} 
The reasoning steps of the AIR reasoner are considered as events and modeled as subclasses of \texttt{air:Event}. \texttt{air:Rule} represents rules, and it is defined as a subclass of \texttt{air:Operation}. The ontology also provides properties to enable representing variable mappings in the performed operations. 
AIR provides a means to write explicit explanations using the assertion property associated with rules. This property is composed of two components, \texttt{air:Statement}, which is the set of triples being asserted, and \texttt{air:Justification}, which is the explicit justification that needs to be associated with the statement.\\ 
\noindent\textbf{Example policy reasoning with explanations using AIR:}\\
Parts of the \emph{Massachusetts Disability Discrimination Law} were translated into a computer interpretable policy using AIR. 
A user's phone records requesting some service and subsequently getting denied based on his disability recorded in the phone logs were captured in RDF. 
Once the AIR reasoner is invoked with the policy file, and the phone log in RDF, a user can visualize the annotated transaction log that contains the reasoning output.
Figure~\ref{chapter01-tabulator_why_pane} contains a partial proof tree with natural language assertions.

\footnotetext{Image available at: \url{http://dig.csail.mit.edu/TAMI/2008/JustificationUI/howto.html}}

\subsubsection{Semantic applications}
Aside from the various representation mechanisms described earlier that support provenance encoding and personalized content, there have been 
many semantic 
applications (e. g., \cite{seneviratne2018knowledge, mccusker2017finding})
enabled by these representations that are explainable. We briefly describe two of these efforts.

In our automatic breast cancer characterization effort \cite{seneviratne2018knowledge}, we developed a visual interface to assist physicians in their diagnosis process by providing justifications of the treatment rules that resulted in a stage change of a patient between changing guideline editions. We considered the 7th and 8th edition of the American Joint Committee on Cancer (AJCC) cancer staging guidelines \cite{giuliano2017breast}. Our system reasoned using
a knowledge base of encoded cancer staging rules, and inferred the stage of the patient based on their metastasis parameters and biomarkers. Our system could automatically determine the staging, explain how 
the stage was derived,  and explain any restaging that happened. In another effort, McCusker et al. developed a framework \cite{mccusker2017finding} that encoded semantic connections between drugs, proteins, and diseases and allowed users to look for potential repurposing of drugs. A novel aspect of this system was that the interface allowed the user to explore why a drug may be used to target a particular disease, thus having a potential causal explanation as opposed to many other drug repurposing efforts that focused only on correlations. The system also included weights on all of the links in the graph so that users could get a sense of how strongly the evidence supports a relationship.

The semantic applications we reviewed primarily utilize scientific evidence to present factual content, discover new content, and automate human-intensive tasks. In Section \ref{chapter01-cogassistants}, we review explanation modeling frameworks, such as, the Inference Web \cite{mcguinness2004explaining}, which also have semantic representations but are used in more typical cognitive assistant settings. 
\subsubsection{Summary}
The Semantic Web efforts we described address various components of explainability. Although, even these interpretable systems, powered by KGs and ontologies, do not entirely address all aspects of explainability that we detail in Section \ref{chapter01-definition}. However, we believe that semantic representations for explainability can evolve from the existing semantic representations for provenance, accountability and context. Hence, we believe that the strengths of the Semantic Web, coupled with ML methods, will be a significant contributor to hybrid explainable AI systems.

\subsection{Cognitive Assistants}
\label{chapter01-cogassistants}
Cognitive Assistants 
are systems that are used to ``augment human intelligence" \cite{engelbart1995toward} and aid humans in decision-making and problem-solving. These assistants have grown from their former role of professional assistants, educating users in a particular domain, to being widely accessible as personal assistants, aiding users in their everyday tasks. These assistants function in a tight coupling with the user and, hence, their design, knowledge bases, and interactions are driven by users' cognitive capabilities and needs. Further, these assistants play various roles from fostering positive behavior change, to training people with the necessary problem-solving skills in a domain, to providing tailored information based on an understanding of user context \cite{ebling2016can}. As the proliferation of general purpose, conversational cognitive assistants grows, it will become increasingly important that they include a representation of the user's goals, and ``theory-of-mind" elements that support effective communication and collaboration. \cite{farrell2016symbiotic}

\subsubsection{DARPA PAL program}
An ambitious and multi-university program, the Defense Advanced Research Projects Agency (DARPA) program, Personal Assistant that Learns (PAL),\footnote{PAL: \url{https://www.darpa.mil/about-us/timeline/personalized-assistant-that-learns}} gave rise to the Cognitive Assistant that Learns and Organizes (CALO) system. CALO was a large effort including over 20 collaboration organizations aimed at building
 a cognitive agent that can assist in a wide range of day-to-day office-related tasks, including sending out emails, memos, maintaining a to-do list \cite{conley2007towel}, etc. Henceforth, several projects
 leveraged the CALO work, the most famous is Apple's personal assistant Siri. In our review, we will cover some of the seminal explainable cognitive assistants \cite{mcguinness2004explaining, mcguinness2007explaining} and user studies \cite{glass2008toward} that resulted from or were refined within the CALO project, that are explainable in their own right. 

Inference Web was one of the early modular explanation frameworks, and it built upon the strengths from the Semantic Web \cite{hendler2002integrating}, Description Logics \cite{baader2004description}, and expert systems communities, to generate explanations for distributed, web-based systems that were interacting with users. The framework provided explanations that contained the provenance of the information (both implicit and explicit), and the proof for inference traces to novice users and agents alike. Additionally, the framework could abstract explanations to suit users' understanding and to avoid lengthy proofs that would overwhelm the users (similar to the breadcrumbs features provided by the CLASSIC system (Section \ref{chapter01-sec:CLASSIC})). 
Besides the ability to abstract explanations, the framework was also capable of providing explanations in different formats and even had a built-in explanation dialogue that would display questions and answers. Users could then interact with the answers and pose follow-up questions. 
The framework achieved its explanation capabilities via a modular architecture consisting of an IWBase, a data repository of the metainformation about the information used by the framework;  an IWAbstractor, abstractor component that converted lengthy Proof Markup Language (PML) \cite{da2006proof} proofs to PML explanations; an IWExplainer, an explanation dialogue component that would generate explanations for users; and an IWBrowser, a browser for displaying the explanations.
While the Inference Web framework did not include a context-specific component, it provided some context modeling options and was capable of providing a wide range of customized explanation capabilities that included direct support for encoding trust and user models.

\begin{figure}[ht!]\centering
\includegraphics[width=1.0\textwidth]{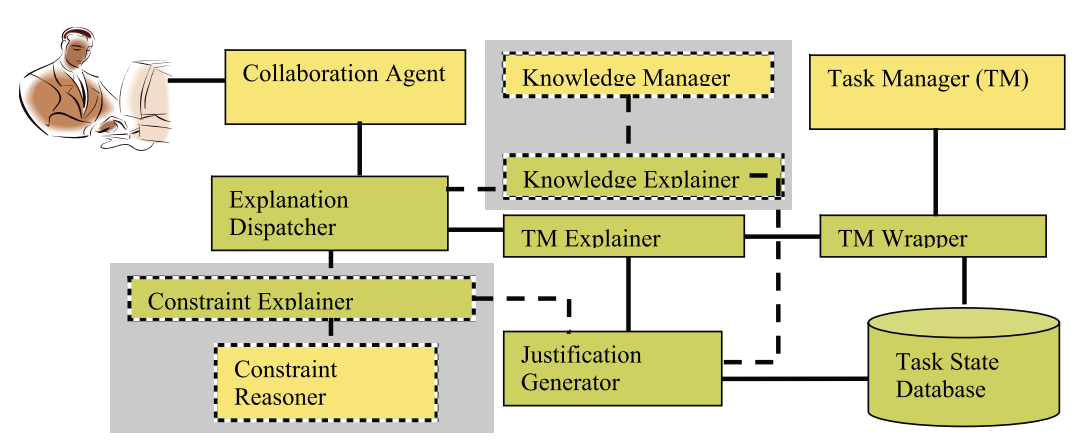}
\caption[icee]{An activity flow diagram of the Integrated Cognitive Explanation Environment (ICEE) that was utilized to explain task-processing systems in the CALO project taken from McGuinness et al.  \cite{mcguinness2007explaining}.}
\label{chapter01-icee}
\end{figure}   

McGuinness et al. \cite{mcguinness2007explaining} expanded on their earlier Inference Web \cite{mcguinness2004explaining} framework, and developed an Integrated Cognitive Explanation Environment (ICEE) that generated explanations for task reasoning. ICEE served as an explanation component on the CALO system, in which multiple reasoning techniques, including task processing, numerous learning components, along with
statistical and deductive methods, 
all worked
together. Since CALO served as a cognitive assistant in the workplace, the tasks involved processing workplace automation activities, such as requesting quotes from different sources (e.g., \textit{GetQuotes} was one of the sub-tasks \cite{mcguinness2007explaining}). 
Additionally, the reasoning techniques used in CALO 
used multiple knowledge sources to generate conclusions that needed to be explained.
The ICEE explanation architecture (shown in Figure \ref{chapter01-icee}) consisted of several components critical to generating explanations: an \textit{explanation dispatcher} that interpreted a user's explanation request and invoked different reasoning components based on the type of explanation request, a \textit{task manager explainer} that further invoked task manager wrappers to gather task execution information, a \textit{task state database} that maintained the execution traces and states of the tasks, and a \textit{justification generator} that created explanations
from the task execution processing information.

The authors conducted a user study aimed at understanding the types of questions that users wanted answered.
These explanation request types included questions about the motivation of a task, status, execution history, forward-looking execution plans, task ordering, or explicit questions about time \cite{mcguinness2007explaining}. 
The classifications of these explanation requests into different request types helped invoke appropriate explanation strategies. Additionally, the system hosted introspective predicates were used to identify the types of information to be included in explanations based on the request's intent. Broadly, the introspective predicates were grouped into \textit{basic procedure information}, metadata about task definitions; \textit{execution information}, details about the task execution; and, \textit{projection information}, information about future task processing. The ICEE framework provides an example of many of the components needed in explainable hybrid AI systems and demonstrated how they can be used to provide user-customized explanations.

Another noteworthy effort from CALO was Glass, et al. \cite{glass2008toward}'s user study that assessed the trust and understandability aspects of adaptive systems. They used the CALO system as an adaptive system use case in their study. Their findings grouped users' concerns into eight themes:
1). \textit{High-level usability of complex prototypes}, 2). \textit{being ignored}, 3). \textit{context-sensitive questions}, 4). \textit{granularity of feedback}, 5). \textit{transparency}, 6). \textit{provenance}, 7). \textit{managing expectations}, and 8). \textit{autonomy and verification}.
While there were some system-related concerns that could be addressed via system improvements (high-level usability, verification), there were also other concerns, such as, provenance, the granularity of feedback, the transparency targeting the users' perception of trust in the agent. They found that the trust level of most users in the system increased significantly with the inclusion of provenance and context-sensitive aspects. Therefore, this study concluded that users who work with cognitive agents would like an interactive dialogue and personalized experience 
and would prefer provenance information to understand the working of these complex systems, to some degree. The themes identified in this paper remain desired features for our complex, hybrid systems of today that use both statistical ML and reasoning techniques.

\subsubsection{Intelligent tutors}
Intelligent tutoring is a sub-domain of cognitive assistants, where adaptive task-oriented systems are utilized for training humans in a particular domain. Hence, intelligent tutors need to appeal to the human cognition and understand and evolve their learning capabilities and grasp of the domain. In a seminal work, VanLeHan \cite{vanlehn2006behavior} noted that there are two loops to human tutoring, an inner and outer loop.  He noted that the inner loop worked in tandem with the human, helping them at each step, assessing their competence, and updating the student model, while the outer loop identified a new task to execute based on the student's assessment. Enhancements have been proposed to VanLeHan's inner and outer loop proposition, one of which is a behavior graph \cite{aleven2009new} that kept track of the possible problem-solving strategies that students can adopt. The edges in a behavior graph represented the different ways in which students could solve problems, and the nodes represented the acceptable states. In general, intelligent tutors host an inherent, domain-specific knowledge component
that is used to undertake tasks.

A use case on explainable, intelligent tutors was explored in a military setting by a
Disciple-LTA \cite{tecuci2005personal} system. They used an iterative problem-solving approach in intelligence tasks to assist analysts. 
These tasks were broken down into executable steps to which evidence could be associated to find solutions (also termed as ``task-reduction"). The solutions were then combined at the task level, or ``solution-composition," to produce conclusions. A sample conclusion from this system was 
``There is strong evidence that Location-A is a training base for terrorist operations." \cite{tecuci2005personal}  The Disciple-LTA architecture consisted of different reasoning agents: learners, tutors, and problem-solvers, all of which read from and wrote into the knowledge base of an ontology and its rules. 

\subsubsection{Summary}
The cognitive assistant literature is vast and continues to grow
with the emergence of personal assistants, such as, Apple's Siri, Amazon's Alexa, etc. In our review, we have covered explanation facilities in DARPA's CALO project \cite{mcguinness2004explaining, mcguinness2007explaining, glass2008toward}, and have also briefly discussed Intelligent Tutors \cite{vanlehn2006behavior, aleven2009new, tecuci2005personal}. While the focus of explanations in the CALO cognitive assistant was  on explaining task-based workflows, the underlying system contained a set of hybrid deductive reasoners coupled with numerous learning components, and thus is representative of today's hybrid learning systems. User requirements were utilized to design explanation strategies and determine the execution of the next task, dictated by user feedback. Cognitive assistants have begun to focus on the end-user, and are supporting facilities to account for user perspective, to some extent, unlike expert systems (Section \ref{chapter01-explainablekbs}) that focused primarily on generating explanations of inference traces.

\subsection{Explainability in  
Machine Learning (ML)
} \label{chapter01-explainable AIworld}
\begin{figure}[ht!]\centering
\includegraphics[width=0.8\textwidth]{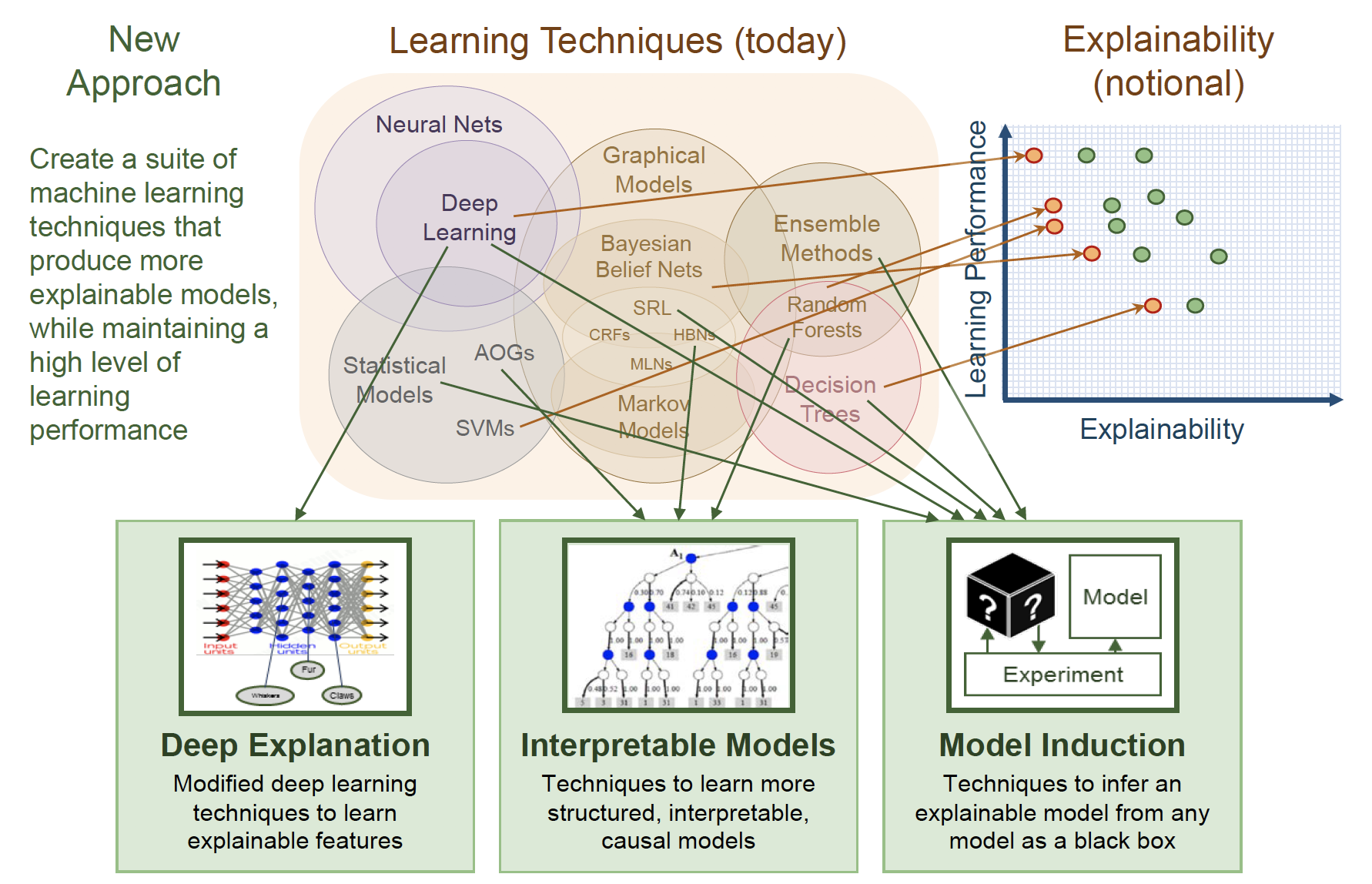}
\caption[darpaexplainable AI]{A high-level overview of the ML models' classes and the explanation techniques being developed as part of DARPA's explainable AI program. It is interesting to note the accuracy-explainability tradeoff depicted in the graph on the right, which shows that within the ML domain, simpler models which are oftentimes less accurate are often more explainable [Image taken from Gunning \cite{gunning2017explainable}].}
\label{chapter01-darpaexplainable AI}
\end{figure}

ML algorithms have been rapidly advancing, proliferating in various domains, even high-precision domains, such as, healthcare and finance. However, these algorithms, are typically more opaque than previous expert systems (Section \ref{chapter01-explainablekbs}), semantic applications (Section \ref{chapter01-semanticweb}), and cognitive assistants (Section \ref{chapter01-cogassistants}).
Hence, the ML domain faces large challenges in addressing the trustworthiness, transparency, and intelligibility\footnote{Our definition of intelligibility is very similar to the description proposed by Lipton \cite{lipton2018mythos} and Lou et al. \cite{lou2012intelligible}, in that intelligible models are interpretable wherein the contribution of model features to a decision can be deciphered.} 
of their models.  Additionally, even within the ML domain, there has been a shift from the dependence on simpler linear algorithms that were less complex, to non-linear, ``black-box" models, such as, deep learning \cite{lecun2015deep}. While ML algorithms are
often
achieving high accuracy, they are 
typically
unable to explain why they arrived at a classification or score (view the tradeoff in Figure \ref{chapter01-darpaexplainable AI}). However, there have been techniques to circumvent these issues, such as, providing confidence scores for the results of models to induce trust (post-hoc interpretations \cite{arya2019one, mittelstadt2019explaining}), attaching semantic information to results \cite{bau2017network}, presenting contrastive or counterfactual explanations to provide intuition for the model's functioning \cite{wachter2017counterfactual, van2018contrastive},  etc. Formally, the interpretability techniques for ML models can be grouped into two categories \cite{mittelstadt2019explaining}, one class aimed at \textit{post-hoc interpretations} that contain explanations about the results to provide perspective on the model's functioning, and the other aimed at improving \textit{transparency} to offer an intuition for the model's functioning. We want to clarify that although ML models might not be considered traditional knowledge-enabled system candidates, we
have included them in our review due to the
emergence of hybrid systems composed of ML models and semantic methods. We believe that a review of explainability approaches in the ML domain will be fruitful for introducing explanation components into these hybrid systems.

\subsubsection{DARPA XAI Program}
DARPA's Explainable AI (XAI) program\footnote{DARPA XAI program website: \url{https://www.darpa.mil/program/explainable-artificial-intelligence}} focuses on building
explainable models that achieve high accuracy and on methods to enable human users to trust and understand these models. 
We will discuss selected XAI efforts mentioned in the DARPA XAI reports \cite{gunning2017explainable, gunning2019darpa} that have a knowledge explainability component to them. 

Bau et. al. \cite{bau2017network}, have developed a \textit{network dissection} technique to align the intermediate layer results of convolutional neural networks (CNN) with semantic concepts. 
They make two 
contributions, a \textit{network dissection} technique to identify what the network is learning at each step by comparing it to semantic concepts, and the construction of \textit{disentangled representation} to align encodings between the network's output and a semantic concept. The disentangled representations were designed to provide a notion of the ``human perception of what it means for a concept to be mixed up" \cite{bau2017network}. Further, the authors also assembled a new dataset, the Broadly and Densely Labeled Dataset (Broden) \cite{zhou2018interpreting} of objects,
that contained low-level compositions of objects used as semantic concepts. This work addresses the deep explanation component of Figure \ref{chapter01-darpaexplainable AI}, wherein feature modifications are being made to make deep learning algorithms interpretable. Similarly, as part of the same program, a team of Charles River Analytica (CRA) researchers developed a technique to learn the causal nature of CNN activations \cite{harradon2018causal}. In this work, Harradon et al. \cite{harradon2018causal} construct a causal graph in-line with Judea Pearl's do-calculus \cite{pearl2017theoretical} method. They ground the network activations in a $P(O, P, C)$ graph, where $C$ represents concepts of network representations that humans can identify, $P$ is the input, and $O$ is the output. However, unlike the network dissection paper \cite{bau2017network}, the causal graph is learned via an unsupervised autoencoder method. Hence, it might be challenging to trust the causal graphs that are learned. 

Since the XAI program by DARPA is an ongoing initiative, some of the work mentioned in the slideware\footnote{DARPA explainable AI slideware: \url{https://www.darpa.mil/attachments/explainable AIProgramUpdate.pdf}} remains unpublished. However, we
briefly summarize some of these unpublished methods
that we believe are relevant to our explainability review.
In the Common Ground Learning and Explanation (COGLE) project,\footnote{COGLE: \url{https://www.parc.com/blog/explainable-ai-an-overview-of-parcs-cogle-project-with-darpa/}} being led by PARC, a system is being built to explain to humans the workings of an autonomous Unmanned Aircraft System (UAS) testbed. 
The COGLE system explains the workings of the UAS reinforcement learning decision-making algorithm to users, conveys an understanding of the system's future behavior, and uses a common ground vocabulary to present these explanations. The common ground vocabulary is generated by including both human understandable and machine-understandable terminology, hence, hoping to ensure a dialogue between ML algorithms and humans. The common ground idea corroborates a requirement put forth by Doshi-Velez et al., \cite{doshi2017accountability} that ``to build AI systems that can provide explanation in terms of human-interpretable terms,
we must both list those terms and allow the AI system access to examples to learn them." In another  
effort, researchers at Rutgers have proposed a technique to choose optimal examples to explain a model's decision via Bayesian Teaching \cite{yang2017explainable}. The explanation by the Bayesian Teaching method is an explanation by examples technique, wherein model-agnostic probabilistic methods are used to identify the most probable data points that lead to a conclusion. The hypothesis that Yang et al. \cite{yang2017explainable} present is that the data is most representative of the algorithm's conclusions, and humans tend to understand more intuitively through examples. 

In summary, the DARPA XAI
program (of which the report is a by-product \cite{gunning2017explainable}) is largely focused on improving explainability of deep learning models through local interpretation methods, or ``knowing the reasons for specific decisions" \cite{doshi2017towards} and post-hoc interpretations. These focus points, to some extent, address the trustworthiness and intelligibility aspects of the explainability of ML models. 

\begin{figure}[ht!]\centering
\includegraphics[width=1.0\textwidth]{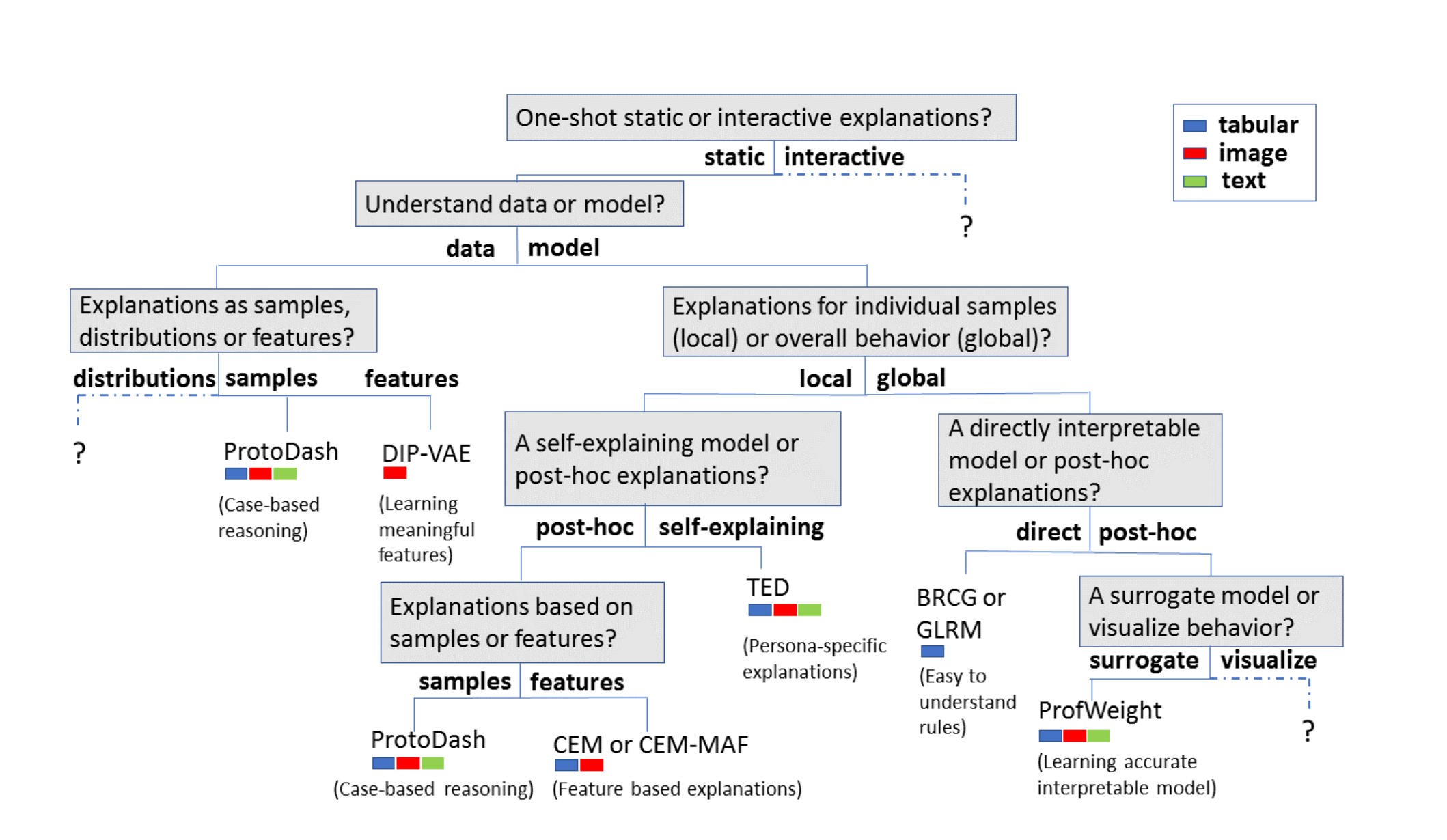}
\caption[explainable AItaxonomy]{A decision-tree like visual overview of the taxonomy of explanations which encodes different factors ML models need to consider while designing explainable models. [Image taken from Arya et al. \cite{arya2019one}]}
\label{explainable AItaxonomy}
\end{figure}

\subsubsection{Taxonomies in explainable ML}
Besides the DARPA XAI program, there have been other recent efforts in the ML domain to support the explainability of ML models. A team of researchers from IBM Research have built the AI Fairness 360 \cite{bellamy2018ai} and AI Explainability 360 \cite{arya2019one} toolkits to identify bias in datasets and ML algorithms, and to describe the explainability of ML models, respectively.
In their AI fairness 360 toolkit, Bellamy et al. \cite{bellamy2018ai} define metrics to identify bias in three stages of the dataset, the algorithm, and the predictions of the algorithm, 
in their goal to improve fairness in the entire ML workflow.
While, in Section \ref{chapter01-introduction}, we noted that we do not account for fairness in explainability, we acknowledge that the exposition of the fairness of the algorithm and data could increase trust in the model. Furthermore, in the AI explainability 360 effort, Arya et al. \cite{arya2019one} designed a taxonomy resource to provide a structure of the explanation space to benefit algorithm designers who are looking to include necessary components in explanations. More specifically, their taxonomy (Figure \ref{explainable AItaxonomy}) helps in identifying methods to introduce local (explanations of portions of the model that lead to a conclusion), global (an explanation of the entire model), and post-hoc interpretations (explaining the results) of models via careful inclusion of features and mechanisms during the design of ML models. However, their taxonomy focused more on model interpretability and features of the model, rather than on intended use of the model by users.

Researchers at MIT conducted a literature review of published explainable AI papers and cataloged the explanation methods used by ML algorithms into a taxonomy \cite{gilpin2018explaining}. Their taxonomy grouped papers into three categories, methods that emulate the processing of the data, explanations of representations (such as, the network dissection technique \cite{bau2017network}), and explanation-producing networks. Their hope was for future methods to use the taxonomy as a reference to build explainable models. Additionally, they note that certain ML methods, such as, decision trees are more interpretable than black-box models and hence are being utilized as proxies \cite{zilke2016deepred} to explain the conclusions of these black-box models. Similarly, Gilpin et al. \cite{gilpin2018explaining} proposed the Local Interpretable Model-agnostic Explanations (LIME) framework \cite{ribeiro2016should}, that can be utilized to generate linear models on perturbations of the black-box model input to get a sense of the functioning of the black-box models.

\subsubsection{Summary}
From a review of the ML domain, we can infer that the explainability techniques being developed are mainly tackling challenges of model interpretability and generating post-hoc interpretations of the model's conclusions or input data. While these two broad categories might seem insufficient, the breadth of innovative approaches \cite{bau2017network, yang2017explainable, arya2019one, wachter2017counterfactual}  being developed are promising and can help in building interpretable, and hybrid models, aided by explainable models (e.g., KGs, causal methods). In summary, what makes the models that we describe in this section candidates for explainable knowledge-enabled systems is that they utilize knowledge to provide an intuition for the functioning of unintelligible models \cite{bau2017network}, or to build a vocabulary (COGLE: \footnote{COGLE: \url{https://www.parc.com/blog/explainable-ai-an-overview-of-parcs-cogle-project-with-darpa/}}) to explain conclusions/inputs/workings of the algorithms. Additionally, prior knowledge of the requirements of explanations are being encoded as taxonomies \cite{arya2019one, gilpin2018explaining} to serve as checks for future explainable models, and knowledge of existing linear models
\cite{zilke2016deepred, ribeiro2016should} are being leveraged to enhance the explanation capabilities of ML models. Orthogonally, the interpretability research in the ML domain is helping researchers understand that humans prefer richer, social, contrasting, and selective explanations \cite{mittelstadt2019explaining}.

\section{Conclusion} \label{chapter01-conclusions} 
We presented foundational approaches to explainable, knowledge-enabled systems, and identified themes for explainability within these approaches.
We presented our definition of ``explainable knowledge-enabled systems" to cover a broad range of past and present AI systems including expert systems, Semantic Web, cognitive assistants, and ML domains. Additionally, we believe that, with the
increasing focus on explainable AI,
we are at the cusp of a new era of AI where explainability plays a pivotal role in the adoption of AI systems.
We provided synthesized, refined definitions of knowledge-enabled systems from a user perspective and included properties that are desirable for when a system needs to generate \textit{provenance-aware, personalized, and context-aware} explanations. 

We reminded our readers that different AI domains and varying methodologies are differently suited for various aspects of explanations.
The next-generation hybrid AI systems would benefit from these identified strengths, utilizing a potentially, carefully chosen 
combination of these techniques 
to provide more complete, satisfying explanations. For instance, we identified that trace-based explanation facilities are well-explored in expert systems, provenance encoding in the Semantic Web domain is capable of representing different granularities of evidence, the modular, task-based explanation facilities of cognitive asssistants can generate atomic explanation components, and that interpretability efforts in the ML domain are giving rise to taxonomical checks for explainable AI models that can be adapted to other AI fields. However, we noted that these AI systems do not fully account for aspects such as user context and causality and are only capable of generating explanations belonging to a restricted set of explanation types. To address these issues, we present directions for research and describe different explanation types in a later chapter, ``Directions for Explainable Knowledge-enabled Systems," that might play a key role in furthering explainable AI. 

In conclusion \footnote{We have used abbreviations in our references as per the IEEE reference guide \url{https://ieeeauthorcenter.ieee.org/wp-content/uploads/IEEE-Reference-Guide.pdf}}, we believe that with the increased adoption of AI systems, there is an increased need for systems to be interpretable, adaptive, interactive, and, most importantly, able to generate explanations that not only provide an overview of the AI system, but serve as a means to educate users and help in their future explorations. To address these lofty goals of explainability, we believe that we need to learn from strengths of past foundational approaches and adapt/expand on them in the user-centric needs of the current AI landscape to build hybrid AI systems that are interpretable, knowledge-enabled, adaptive, and context and provenance-aware.

\section{Acknowledgments}
This work is partially supported by IBM Research AI through the AI Horizons Network. We thank our colleagues from IBM Research, Amar Das, Morgan Foreman and Ching-Hua Chen, and from RPI, James P. McCusker, and Rebecca Cowan, who greatly assisted the research and document preparation.
\makeatother
\bibliographystyle{IEEETrans}
\bibliography{1-charigruensenevomcguinness}

\begin{thebibliography}{10}
\providecommand{\url}[1]{#1}
\csname url@samestyle\endcsname
\providecommand{\newblock}{\relax}
\providecommand{\bibinfo}[2]{#2}
\providecommand{\BIBentrySTDinterwordspacing}{\spaceskip=0pt\relax}
\providecommand{\BIBentryALTinterwordstretchfactor}{4}
\providecommand{\BIBentryALTinterwordspacing}{\spaceskip=\fontdimen2\font plus
\BIBentryALTinterwordstretchfactor\fontdimen3\font minus
  \fontdimen4\font\relax}
\providecommand{\BIBforeignlanguage}[2]{{%
\expandafter\ifx\csname l@#1\endcsname\relax
\typeout{** WARNING: IEEEtran.bst: No hyphenation pattern has been}%
\typeout{** loaded for the language `#1'. Using the pattern for}%
\typeout{** the default language instead.}%
\else
\language=\csname l@#1\endcsname
\fi
#2}}
\providecommand{\BIBdecl}{\relax}
\BIBdecl

\bibitem{norman1988psychology}
D.~A. Norman, \emph{The psychology of everyday things.}\hskip 1em plus 0.5em
  minus 0.4em\relax Basic books, 1988.

\bibitem{hollan2000distributed}
J.~Hollan, E.~Hutchins, and D.~Kirsh, ``Distributed cognition: toward a new
  foundation for human-computer interaction research,'' \emph{ACM Transactions
  on Computer-Human Interaction (TOCHI)}, vol.~7, no.~2, pp. 174--196, 2000.

\bibitem{hutchins1990technology}
E.~Hutchins, ``The technology of team navigation, Intellectual teamwork: social
  and technological foundations of cooperative work, L,'' 1990.

\bibitem{swartout1991explanations}
W.~Swartout, C.~Paris, and J.~Moore, ``Explanations in knowledge systems:
  Design for explainable expert systems,'' \emph{IEEE Expert}, vol.~6, no.~3,
  pp. 58--64, 1991.

\bibitem{shortliffe1974mycin}
E.~H. Shortliffe, ``MYCIN: a rule-based computer program for advising
  physicians regarding antimicrobial therapy selection.'' Dept. of Computer
  Sci., Stanford University Stanford, Tech. Rep., 1974.

\bibitem{clancey1982neomycin}
W.~J. Clancey and R.~Letsinger, \emph{NEOMYCIN: Reconfiguring a rule-based
  expert system for application to teaching}.\hskip 1em plus 0.5em minus
  0.4em\relax Dept. of Computer Sci., Stanford University Stanford, 1982.

\bibitem{mcguinness2007explaining}
D.~L. McGuinness, A.~Glass, M.~Wolverton, and P.~P. Da~Silva, ``Explaining Task
  Processing in Cognitive Assistants That Learn.'' in \emph{AAAI Spring Symp.:
  Interaction Challenges for Intelligent Assistants}, 2007, pp. 80--87.

\bibitem{seneviratne2018knowledge}
O.~Seneviratne, S.~M. Rashid, S.~Chari, J.~P. McCusker, K.~P. Bennett, J.~A.
  Hendler, and D.~L. McGuinness, ``Knowledge Integration for Disease
  Characterization: A Breast Cancer Example,'' in \emph{Int. Semantic Web
  Conf.}\hskip 1em plus 0.5em minus 0.4em\relax San Francisco, USA: Springer,
  2018, pp. 223--238.

\bibitem{gunning2017explainable}
D.~Gunning, ``Explainable artificial intelligence (XAI),'' \emph{Defense
  Advanced Research Projects Agency (DARPA), nd Web}, vol.~2, 2017.

\bibitem{hendler2002integrating}
J.~Hendler and E.~Miller, ``Integrating applications on the semantic web,''
  \emph{J. of the Institute of Electrical Engineers of Japan}, vol. Vol
  122(10), pp. 676--680, 2002.

\bibitem{lecun2015deep}
Y.~LeCun, Y.~Bengio, and G.~Hinton, ``Deep learning,'' \emph{Nature}, vol. 521,
  no. 7553, pp. 436--444, 2015.

\bibitem{dhaliwal1996use}
J.~S. Dhaliwal and I.~Benbasat, ``The use and effects of knowledge-based system
  explanations: theoretical foundations and a framework for empirical
  evaluation,'' \emph{Information systems research}, vol.~7, no.~3, pp.
  342--362, 1996.

\bibitem{shortliffe2012computer}
E.~Shortliffe, \emph{Computer-based medical consultations: MYCIN}.\hskip 1em
  plus 0.5em minus 0.4em\relax Elsevier, 2012.

\bibitem{lou2012intelligible}
Y.~Lou, R.~Caruana, and J.~Gehrke, ``Intelligible models for classification and
  regression,'' in \emph{Proc. of the 18th ACM SIGKDD Int. Conf. on Knowledge
  discovery and data mining}.\hskip 1em plus 0.5em minus 0.4em\relax ACM, 2012,
  pp. 150--158.

\bibitem{holzinger2017we}
A.~Holzinger, C.~Biemann, C.~S. Pattichis, and D.~B. Kell, ``What do we need to
  build explainable AI systems for the medical domain?'' \emph{arXiv preprint
  arXiv:1712.09923}, 2017.

\bibitem{mcguinness2004owl}
D.~L. McGuinness, F.~Van~Harmelen \emph{et~al.}, ``OWL web ontology language
  overview,'' \emph{W3C recommendation}, vol.~10, no.~10, p. 2004, 2004.

\bibitem{lebo2013prov}
T.~Lebo, S.~Sahoo, D.~McGuinness, K.~Belhajjame, J.~Cheney, D.~Corsar,
  D.~Garijo, S.~Soiland-Reyes, S.~Zednik, and J.~Zhao, ``PROV-O: The prov
  ontology,'' \emph{W3C recommendation}, 2013.

\bibitem{groth2010anatomy}
P.~Groth, A.~Gibson, and J.~Velterop, ``The anatomy of a nanopublication,''
  \emph{Information Services \& Use}, vol.~30, no. 1-2, pp. 51--56, 2010.

\bibitem{mittelstadt2019explaining}
B.~Mittelstadt, C.~Russell, and S.~Wachter, ``Explaining explanations in AI,''
  in \emph{Proc. of the Conf. on fairness, accountability, and
  transparency}.\hskip 1em plus 0.5em minus 0.4em\relax ACM, 2019, pp.
  279--288.

\bibitem{biran2017explanation}
O.~Biran and C.~Cotton, ``Explanation and justification in machine learning: A
  survey,'' in \emph{IJCAI-17 workshop on explainable AI (XAI)}, vol.~8, 2017,
  p.~1.

\bibitem{arya2019one}
V.~Arya, R.~K. Bellamy, P.-Y. Chen, A.~Dhurandhar, M.~Hind, S.~C. Hoffman,
  S.~Houde, Q.~V. Liao, R.~Luss, A.~Mojsilovi{\'c} \emph{et~al.}, ``One
  Explanation Does Not Fit All: A Toolkit and Taxonomy of AI Explainability
  Techniques,'' \emph{arXiv preprint arXiv:1909.03012}, 2019.

\bibitem{swartout1993explanation}
W.~R. Swartout and J.~D. Moore, ``Explanation in second generation expert
  systems,'' in \emph{Second generation expert systems}.\hskip 1em plus 0.5em
  minus 0.4em\relax Springer, 1993, pp. 543--585.

\bibitem{doshi2017accountability}
F.~Doshi-Velez, M.~Kortz, R.~Budish, C.~Bavitz, S.~Gershman, D.~O'Brien,
  S.~Schieber, J.~Waldo, D.~Weinberger, and A.~Wood, ``{\relax Accountability
  of AI under the law: The role of explanation},'' \emph{arXiv preprint
  arXiv:1711.01134}, 2017.

\bibitem{gilpin2018explaining}
L.~H. Gilpin, D.~Bau, B.~Z. Yuan, A.~Bajwa, M.~Specter, and L.~Kagal,
  ``Explaining explanations: An approach to evaluating interpretability of
  machine learning,'' \emph{arXiv preprint arXiv:1806.00069}, 2018.

\bibitem{miller2019explanation}
T.~Miller, ``Explanation in artificial intelligence: Insights from the social
  sciences,'' \emph{Artificial Intelligence}, vol. 267, pp. 1--38, 2019.

\bibitem{lim2009and}
B.~Y. Lim, A.~K. Dey, and D.~Avrahami, ``Why and why not explanations improve
  the intelligibility of context-aware intelligent systems,'' in \emph{Proc. of
  the SIGCHI Conf. on Human Factors in Computing Systems}.\hskip 1em plus 0.5em
  minus 0.4em\relax ACM, 2009, pp. 2119--2128.

\bibitem{ribera2019can}
M.~Ribera and {\`A}.~Lapedriza, ``Can we do better explanations? A proposal of
  user-centered explainable AI.'' in \emph{IUI Workshops}, 2019.

\bibitem{maimone2018perkapp}
R.~Maimone, M.~Guerini, M.~Dragoni, T.~Bailoni, and C.~Eccher, ``PerKApp: A
  general purpose persuasion architecture for healthy lifestyles,'' \emph{J. of
  biomedical informatics}, vol.~82, pp. 70--87, 2018.

\bibitem{glass2008toward}
A.~Glass, D.~L. McGuinness, and M.~Wolverton, ``Toward establishing trust in
  adaptive agents,'' in \emph{Proc. of the 13th Int. Conf. on Intelligent user
  interfaces}.\hskip 1em plus 0.5em minus 0.4em\relax ACM, 2008, pp. 227--236.

\bibitem{rich1989stereotypes}
E.~Rich, ``Stereotypes and user modeling,'' in \emph{User models in dialog
  systems}.\hskip 1em plus 0.5em minus 0.4em\relax Springer, 1989, pp. 35--51.

\bibitem{sugiyama2004adaptive}
K.~Sugiyama, K.~Hatano, and M.~Yoshikawa, ``Adaptive web search based on user
  profile constructed without any effort from users,'' in \emph{Proc. of the
  13th Int. Conf. on World Wide Web}.\hskip 1em plus 0.5em minus 0.4em\relax
  ACM, 2004, pp. 675--684.

\bibitem{doshi2017towards}
F.~Doshi-Velez and B.~Kim, ``Towards a rigorous science of interpretable
  machine learning,'' \emph{arXiv preprint arXiv:1702.08608}, 2017.

\bibitem{air-policy-language}
\BIBentryALTinterwordspacing
L.~Kagal, ``{Accountability In RDF (AIR) Web Rule Language},'' 2009. [Online].
  Available: \url{http://dig.csail.mit.edu/2009/AIR/}
\BIBentrySTDinterwordspacing

\bibitem{pearl2009causality}
J.~Pearl, \emph{Causality}.\hskip 1em plus 0.5em minus 0.4em\relax Cambridge
  university press, 2009.

\bibitem{mcguinness2004explaining}
D.~L. McGuinness and P.~P. Da~Silva, ``Explaining answers from the semantic
  web: The inference web approach,'' \emph{Web Semantics: Sci., Services and
  Agents on the World Wide Web}, vol.~1, no.~4, pp. 397--413, 2004.

\bibitem{lecue2019role}
F.~Lecue, ``On the role of knowledge graphs in explainable AI,'' \emph{Semantic
  Web J. (Forthcoming)}, 2019.

\bibitem{hasan2012explanation}
R.~Hasan and F.~Gandon, ``Explanation in the Semantic Web: a survey of the
  state of the art,'' Research. Rep., 2012.

\bibitem{borgida1989classic}
A.~Borgida, R.~J. Brachman, D.~L. McGuinness, and L.~A. Resnick, ``CLASSIC: A
  structural data model for objects,'' in \emph{ACM Sigmod record}, vol.~18,
  no.~2.\hskip 1em plus 0.5em minus 0.4em\relax ACM, 1989, pp. 58--67.

\bibitem{tecuci2005personal}
G.~Tecuci, M.~Boicu, C.~Ayers, and D.~Cammons, ``Personal cognitive assistants
  for military intelligence analysis: Mixed-initiative learning, tutoring, and
  problem solving,'' in \emph{First Int. Conf. on Intelligence Analysis}.\hskip
  1em plus 0.5em minus 0.4em\relax Citeseer, 2005, pp. 2--6.

\bibitem{ribeiro2016should}
M.~T. Ribeiro, S.~Singh, and C.~Guestrin, ``Why should i trust you?: Explaining
  the predictions of any classifier,'' in \emph{Proc. of the 22nd ACM SIGKDD
  Int. Conf. on knowledge discovery and data mining}.\hskip 1em plus 0.5em
  minus 0.4em\relax ACM, 2016, pp. 1135--1144.

\bibitem{mccusker2017finding}
J.~P. McCusker, M.~Dumontier, R.~Yan, S.~He, J.~S. Dordick, and D.~L.
  McGuinness, ``Finding melanoma drugs through a probabilistic knowledge
  graph,'' \emph{PeerJ Computer Sci.}, vol.~3, p. e106, 2017.

\bibitem{gunning2019darpa}
D.~Gunning and D.~W. Aha, ``DARPA's Explainable Artificial Intelligence
  Program,'' \emph{AI Magazine}, vol.~40, no.~2, pp. 44--58, 2019.

\bibitem{patel1991classic}
P.~F. Patel-Schneider, D.~L. McGuinness, R.~J. Brachman, and L.~A. Resnick,
  ``The CLASSIC knowledge representation system: Guiding principles and
  implementation rationale,'' \emph{ACM SIGART Bulletin}, vol.~2, no.~3, pp.
  108--113, 1991.

\bibitem{mccusker2017broad}
J.~P. McCusker, S.~M. Rashid, Z.~Liang, Y.~Liu, K.~Chastain, P.~Pinheiro, J.~A.
  Stingone, and D.~L. McGuinness, ``Broad, Interdisciplinary Science In Tela:
  An Exposure and Child Health Ontology,'' in \emph{Proc. of the 2017 ACM on
  Web Sci. Conf.}\hskip 1em plus 0.5em minus 0.4em\relax ACM, 2017, pp.
  349--357.

\bibitem{nunes2017systematic}
I.~Nunes and D.~Jannach, ``A systematic review and taxonomy of explanations in
  decision support and recommender systems,'' \emph{User Modeling and
  User-Adapted Interaction}, vol.~27, no. 3-5, pp. 393--444, 2017.

\bibitem{mcguinness1995explaining}
D.~L. McGuinness and A.~Borgida, ``Explaining subsumption in description
  logics,'' in \emph{IJCAI (1)}, 1995, pp. 816--821.

\bibitem{mcguinness1996explaining}
D.~L. McGuinness, ``Explaining reasoning in description logics,'' Ph.D.
  dissertation, Rutgers University, 1996.

\bibitem{brachman1989overview}
R.~J. Brachman and J.~G. Schmolze, ``An overview of the KL-ONE knowledge
  representation system,'' in \emph{Readings in artificial intelligence and
  databases}.\hskip 1em plus 0.5em minus 0.4em\relax Elsevier, 1989, pp.
  207--230.

\bibitem{berners1994world}
T.~Berners-Lee, D.~Dimitroyannis, A.~J. Mallinckrodt, and S.~McKay, ``World
  Wide Web,'' \emph{Computers in Physics}, vol.~8, no.~3, pp. 298--299, 1994.

\bibitem{berners2001semantic}
T.~Berners-Lee, J.~Hendler, O.~Lassila \emph{et~al.}, ``The semantic web,''
  \emph{Scientific american}, vol. 284, no.~5, pp. 28--37, 2001.

\bibitem{hogan2009scalable}
A.~Hogan, A.~Harth, and A.~Polleres, ``Scalable authoritative OWL reasoning for
  the web,'' \emph{Int. J. on Semantic Web and Information Systems (IJSWIS)},
  vol.~5, no.~2, 2009.

\bibitem{ehrlinger2016towards}
L.~Ehrlinger and W.~W{\"o}{\ss}, ``Towards a Definition of Knowledge Graphs.''
  \emph{SEMANTiCS (Posters, Demos, SuCCESS)}, vol.~48, 2016.

\bibitem{kuhn2013broadening}
T.~Kuhn, P.~E. Barbano, M.~L. Nagy, and M.~Krauthammer, ``Broadening the scope
  of nanopublications,'' in \emph{Extended Semantic Web Conf.}\hskip 1em plus
  0.5em minus 0.4em\relax Springer, 2013.

\bibitem{valdez2017provcare}
J.~Valdez, M.~Kim, M.~Rueschman, V.~Socrates, S.~Redline, and S.~S. Sahoo,
  ``ProvCaRe semantic provenance knowledgebase: evaluating scientific
  reproducibility of research studies,'' in \emph{AMIA Annual Symp. Proc.},
  vol. 2017.\hskip 1em plus 0.5em minus 0.4em\relax American Medical
  Informatics Association, 2017, p. 1705.

\bibitem{agun2019gprov}
N.~N. Agu, N.~Keshan, S.~Chari, O.~Seneviratne, J.~P. McCusker, A.~Das, and
  D.~L. McGuinness, ``G-PROV: Provenance Management for Clinical Practice
  Guidelines,'' in \emph{Proc. of the Semantic Web solutions for large-scale
  biomedical data analytics Workshop}.\hskip 1em plus 0.5em minus 0.4em\relax
  CEUR, 2019, p. to appear.

\bibitem{hertling2017webisalod}
S.~Hertling and H.~Paulheim, ``WebIsALOD: providing hypernymy relations
  extracted from the web as linked open data,'' in \emph{Int. Semantic Web
  Conf.}\hskip 1em plus 0.5em minus 0.4em\relax Springer, 2017, pp. 111--119.

\bibitem{gyrard2018personalized}
A.~Gyrard, M.~Gaur, S.~Shekarpour, K.~Thirunarayan, and A.~Sheth,
  ``Personalized Health Knowledge Graph,'' in \emph{Int. Semantic Web Conf.
  (ISWC) 2018 Contextualized Knowledge Graph Workshop}, 2018.

\bibitem{horridge2008laconic}
M.~Horridge, B.~Parsia, and U.~Sattler, ``Laconic and precise justifications in
  OWL,'' in \emph{Int. semantic web Conf.}\hskip 1em plus 0.5em minus
  0.4em\relax Springer, 2008, pp. 323--338.

\bibitem{kleer1977amord}
J.~d. Kleer, J.~Doyle, G.~L. Steele~Jr, and G.~J. Sussman, ``{AMORD explicit
  control of reasoning},'' \emph{ACM SIGPLAN Notices}, vol.~12, no.~8, pp.
  116--125, 1977.

\bibitem{berners2008n3logic}
T.~Berners-Lee, D.~Connolly, L.~Kagal, Y.~Scharf, and J.~Hendler, ``N3logic: A
  logical framework for the world wide web,'' \emph{Theory and Practice of
  Logic Programming}, vol.~8, no.~3, pp. 249--269, 2008.

\bibitem{giuliano2017breast}
A.~E. Giuliano, J.~L. Connolly, S.~B. Edge, E.~A. Mittendorf, H.~S. Rugo, L.~J.
  Solin, D.~L. Weaver, D.~J. Winchester, and G.~N. Hortobagyi, ``{\relax Breast
  cancer—major changes in the American Joint Committee on Cancer eighth
  edition cancer staging manual},'' \emph{CA: A Cancer J. for Clinicians},
  vol.~67, no.~4, pp. 290--303, 2017.

\bibitem{engelbart1995toward}
D.~C. Engelbart, ``Toward augmenting the human intellect and boosting our
  collective IQ,'' \emph{Communications of the ACM}, vol.~38, no.~8, pp.
  30--33, 1995.

\bibitem{ebling2016can}
M.~R. Ebling, ``Can cognitive assistants disappear?'' \emph{IEEE Pervasive
  Computing}, vol.~15, no.~3, pp. 4--6, 2016.

\bibitem{farrell2016symbiotic}
R.~G. Farrell, J.~Lenchner, J.~O. Kephjart, A.~M. Webb, M.~J. Muller, T.~D.
  Erikson, D.~O. Melville, R.~K. Bellamy, D.~M. Gruen, J.~H. Connell
  \emph{et~al.}, ``Symbiotic cognitive computing,'' \emph{AI Magazine},
  vol.~37, no.~3, pp. 81--93, 2016.

\bibitem{conley2007towel}
K.~Conley and J.~Carpenter, ``Towel: Towards an Intelligent To-Do List.'' in
  \emph{AAAI Spring Symp.: Interaction Challenges for Intelligent Assistants},
  2007.

\bibitem{baader2004description}
F.~Baader, I.~Horrocks, and U.~Sattler, ``Description logics,'' in
  \emph{Handbook on ontologies}.\hskip 1em plus 0.5em minus 0.4em\relax
  Springer, 2004, pp. 3--28.

\bibitem{da2006proof}
P.~P. Da~Silva, D.~L. McGuinness, and R.~Fikes, ``A proof markup language for
  semantic web services,'' \emph{Information Systems}, vol.~31, no. 4-5, pp.
  381--395, 2006.

\bibitem{vanlehn2006behavior}
K.~Vanlehn, ``The behavior of tutoring systems,'' \emph{Int. J.of artificial
  intelligence in education}, vol.~16, no.~3, pp. 227--265, 2006.

\bibitem{aleven2009new}
V.~Aleven, B.~M. Mclaren, J.~Sewall, and K.~R. Koedinger, ``A new paradigm for
  intelligent tutoring systems: Example-tracing tutors,'' \emph{Int. J. of
  Artificial Intelligence in Education}, vol.~19, no.~2, pp. 105--154, 2009.

\bibitem{lipton2018mythos}
Z.~C. Lipton, ``The mythos of model interpretability,'' \emph{Queue}, vol.~16,
  no.~3, pp. 31--57, 2018.

\bibitem{bau2017network}
D.~Bau, B.~Zhou, A.~Khosla, A.~Oliva, and A.~Torralba, ``Network dissection:
  Quantifying interpretability of deep visual representations,'' in \emph{Proc.
  of the IEEE Conf. on Computer Vision and Pattern Recognition}, 2017, pp.
  6541--6549.

\bibitem{wachter2017counterfactual}
S.~Wachter, B.~Mittelstadt, and C.~Russell, ``Counterfactual Explanations
  without Opening the Black Box: Automated Decisions and the GPDR,''
  \emph{Harv. JL \& Tech.}, vol.~31, p. 841, 2017.

\bibitem{van2018contrastive}
J.~van~der Waa, M.~Robeer, J.~van Diggelen, M.~Brinkhuis, and M.~Neerincx,
  ``Contrastive explanations with local foil trees,'' \emph{arXiv preprint
  arXiv:1806.07470}, 2018.

\bibitem{zhou2018interpreting}
B.~Zhou, D.~Bau, A.~Oliva, and A.~Torralba, ``{\relax Interpreting deep visual
  representations via network dissection},'' \emph{IEEE transactions on pattern
  analysis and machine intelligence}, 2018.

\bibitem{harradon2018causal}
M.~Harradon, J.~Druce, and B.~Ruttenberg, ``Causal learning and explanation of
  deep neural networks via autoencoded activations,'' \emph{arXiv preprint
  arXiv:1802.00541}, 2018.

\bibitem{pearl2017theoretical}
J.~Pearl, ``Theoretical impediments to machine learning,'' 2017.

\bibitem{yang2017explainable}
S.~C.-H. Yang and P.~Shafto, ``Explainable Artificial Intelligence via Bayesian
  Teaching,'' in \emph{NIPS 2017 workshop on Teaching Machines, Robots, and
  Humans}, 2017.

\bibitem{bellamy2018ai}
R.~K. Bellamy, K.~Dey, M.~Hind, S.~C. Hoffman, S.~Houde, K.~Kannan, P.~Lohia,
  J.~Martino, S.~Mehta, A.~Mojsilovic \emph{et~al.}, ``AI fairness 360: An
  extensible toolkit for detecting, understanding, and mitigating unwanted
  algorithmic bias,'' \emph{arXiv preprint arXiv:1810.01943}, 2018.

\bibitem{zilke2016deepred}
J.~R. Zilke, E.~L. Menc{\'\i}a, and F.~Janssen, ``DeepRED--Rule extraction from
  deep neural networks,'' in \emph{Int. Conf. on Discovery Sci.}\hskip 1em plus
  0.5em minus 0.4em\relax Springer, 2016, pp. 457--473.

\end{thebibliography}

\end{document}